\documentclass[sigconf, screen]{acmart}

\AtBeginDocument{%
  }

\setcopyright{none} 
\copyrightyear{2026}
\acmYear{2026}
\acmDOI{XXXXXXX.XXXXXXX}

\acmConference[MICRO 2026]{The 58th IEEE/ACM International Symposium on Microarchitecture}{October 31--November 04, 2026}{Athens, Greece}
\acmISBN{978-X-XXXX-XXXX-X/XX/XX}

\usepackage{amsmath,amsfonts}
\usepackage{algorithmic}
\usepackage{graphicx}
\usepackage{textcomp}
\usepackage{xcolor}
\usepackage{subfigure}
\usepackage{multirow}
\usepackage{tikz}
\usepackage[ruled,vlined,linesnumbered]{algorithm2e}
\usepackage{pifont}

\begin{document}

\title{Algorithm–Architecture Co-Design for Efficient VLA Inference via Speculative Inference and Verification}
% \subtitle{\normalsize{MICRO 2026 Submission
%     \textbf{\#1493} -- Confidential Draft -- Do NOT Distribute!!}}

\author{Chunyu Qi$^{1}$, Zhuoran Song$^{1}$, Jian Weng$^{2}$, Haozhe Jiang$^{1}$, Xueyuan Liu$^{1}$,
Naifeng Jing$^{1}$, \\ Guanghui He$^{1}$, Xiaoyao Liang$^{1}$, Haibing Guan$^{1}$}
% \affiliation{%
%     \institution{
%     $^{1}$School of Computer Science, Shanghai Jiao Tong University, Shanghai, China\\
%     $^{2}$Computer Science, King Abdullah University of Science and Technology, Thuwal, Mecca, Saudi Arabia\\}
% }
\affiliation{%
    \institution{
    $^{1}$School of Computer Science, Shanghai Jiao Tong University}
    \city{Shanghai}
    \country{China}
}
\affiliation{%
    \institution{
    $^{2}$Computer Science, King Abdullah University of Science and Technology}
    \city{Thuwal, Mecca}
    \country{Saudi Arabia}
}
\renewcommand{\shortauthors}{Chunyu Qi and Zhuoran Song, et al.}

%\author{\normalsize{MICRO 2026 Submission
 %   \textbf{\#NaN} -- Confidential Draft -- Do NOT Distribute!!}}



%%%%%% -- PAPER CONTENT STARTS-- %%%%%%%%

\begin{abstract}

Vision-Language-Action (VLA) models have demonstrated remarkable capabilities in the field of embodied AI, but their high computational cost and limited predicted action length hinder real-time deployment. Although Dadu-Corki, a dedicated accelerator for efficient embodied AI, has been introduced, it does not exploit the inherent interaction patterns between the robot and its environment, which results in a relatively short predicted action length. We observe that robotic environments naturally alternate between active states—where precise actions are crucial—and inactive states—where actions have limited impact on task success. This insight enables a new scheduling opportunity: long-action-length speculative prediction in inactive states, paired with selective verification in active states.

We propose SpecVLA, an algorithm–system co-design framework that adaptively balances action length, inference latency, and task reliability. On the algorithm side, SpecVLA introduces a state-aware VLA inference execution paradigm and a hardware-friendly construction of a smaller verification model (sVLA) using differential residuals and block-wise mixed-precision quantization. On the system side, we develop a heterogeneous architecture consisting of a GPU and a robotic-specific hardware module, along with a speculative dataflow that decouples VLA and sVLA through parallel execution. The hardware module integrates preprocessing, state prediction, and low-precision compute units into a single tightly coupled pipeline, eliminating redundant data movement and enabling low-latency verification. Comprehensive evaluations on OpenVLA and RDT across LIBERO and ManiSkill benchmarks show that SpecVLA reduces end-to-end latency significantly while preserving task success rate. By enabling long-action-length speculative prediction with timely verification, SpecVLA achieves real-time robotic manipulation with both high efficiency and reliability.

\end{abstract}
%\begin{CCSXML}
%<ccs2012>
%  <concept_id>00000000.0000000.0000000</concept_id>
%  <concept_desc>Do Not Use This Code, Generate the Correct Terms for Your Paper</concept_desc>
%  <concept_significance>500</concept_significance>
%  %<concept_id>00000000.00000000.00000000</concept_id>
%  <concept_desc>Do Not Use This Code, Generate the Correct Terms for Your Paper</concept_desc>
%  <concept_significance>300</concept_significance>
%  %<concept_id>00000000.00000000.00000000</concept_id>
%  <concept_desc>Do Not Use This Code, Generate the Correct Terms for Your Paper</concept_desc>
%  <concept_significance>100</concept_significance>
 % <concept_id>00000000.00000000.00000000</concept_id>
%  <concept_desc>Do Not Use This Code, Generate the Correct Terms for Your Paper</concept_desc>
%  <concept_significance>100</concept_significance>
%</ccs2012>
%\end{CCSXML}

%\ccsdesc[500]{Do Not Use This Code~Generate the Correct Terms for Your Paper}
%\ccsdesc[300]{Do Not Use This Code~Generate the Correct Terms for Your Paper}
%\ccsdesc{Do Not Use This Code~Generate the Correct Terms for Your Paper}
%\ccsdesc[100]{Do Not Use This Code~Generate the Correct Terms for Your Paper}


\maketitle

\section{Introduction}
Vision-Language-Action (VLA) models have demonstrated strong comprehension and generalization capabilities in embodied AI tasks by predicting action sequences based on visual images and language instructions. However, current VLAs suffer from inference latencies exceeding 600 ms, which is too long for real-time control systems. For instance, although autonomous robots can leverage VLAs to support construction automation~\cite{Wang_2025}, like transporting or installing building materials to designated locations, delayed action predictions lead to reduced productivity and increased costs.

To realize efficient embodied AI tasks, Dadu-Corki~\cite{DBLP:conf/isca/HuangH0Y0M00L0G25} proposes a heterogeneous system, where computing-intensive VLA inference is deployed on a GPU server while robot actions are executed on an FPGA platform. To hide the communication and inference latency, Dadu-Corki streams captured environmental frames to the GPU server when the robot is executing actions. Yet, Dadu-Corki overlooks the opportunities of leveraging the continuity of the captured environmental frames, so it can only predict fewer than four actions per inference, leaving substantial unexploited performance.

%the primary performance bottleneck of Dadu-Corki remains VLA inference, restricting its ability to achieve optimal performance gains.

%Given that the inefficiency of VLAs primarily arises from two factors: 1) the high latency of each VLA inference, and 2) the limited number of actions that are predicted per VLA inference. 

To take advantage of the continuity of the robotic environment, we have two key observations: 1. Not all actions contribute equally to task success. We characterize the action execution into \emph{active and inactive states}. \emph{Active states} occur when the robot is carefully manipulating an object (e.g., aligning, grasping, or placing), where precise, fine-grained control is required. \emph{Inactive states} occur when the robot is merely moving the gripper toward a target pose, where the intermediate actions have a limited effect on the final outcome. This asymmetry offers an opportunity for self-adaptive prediction, aggressively predicting long action sequences during inactive states to amortize VLA inference latency, and predicting shorter sequences during active states for the task success rate. 2. The consecutive environment frames exhibit high temporal redundancy. This suggests that a quantized \underline{s}maller VLA (sVLA) variant to the original VLA can still produce high-quality action predictions with significantly lower computing cost and latency. Our evaluation shows that this quantized small model achieves comparable performance for short sequences, and degrades badly for long sequences.

\if 0
A key insight underlying this work is that robotic manipulation tasks naturally alternate between \textbf{inactive} states, where the gripper travels toward a target and actions have minimal effect on task success, and \textbf{active} states, where the gripper interacts with the target and precise action control becomes crucial. This environmental asymmetry exposes an opportunity unexploited by prior systems: \emph{inactive states tolerate long-action-length speculative predictions, whereas active states require short-action-length with reliable verification.}

Moreover, due to the mechanical characteristics of robot arms, sensor observations collected in consecutive frames exhibit strong temporal redundancy. By subtracting consecutive observations, the resulting residual features show a much narrower value distribution compared to raw features, suggesting the feasibility of constructing a lightweight VLA variant (sVLA) using differential-based quantization. While sVLA runs significantly faster than VLA, its accuracy degrades for long-action-length predictions, limiting its ability to fully replace VLA.
\fi

%To achieve the best of both worlds,amortized inference latency for long sequences, and high task success rate, by taking advantage of the observations above, we present SpecVLA, an algorithm–system co-designed framework for real-time action predictions with adaptive sequence length.
\begin{figure*}[h]
\centering
\includegraphics[width=0.8\linewidth]{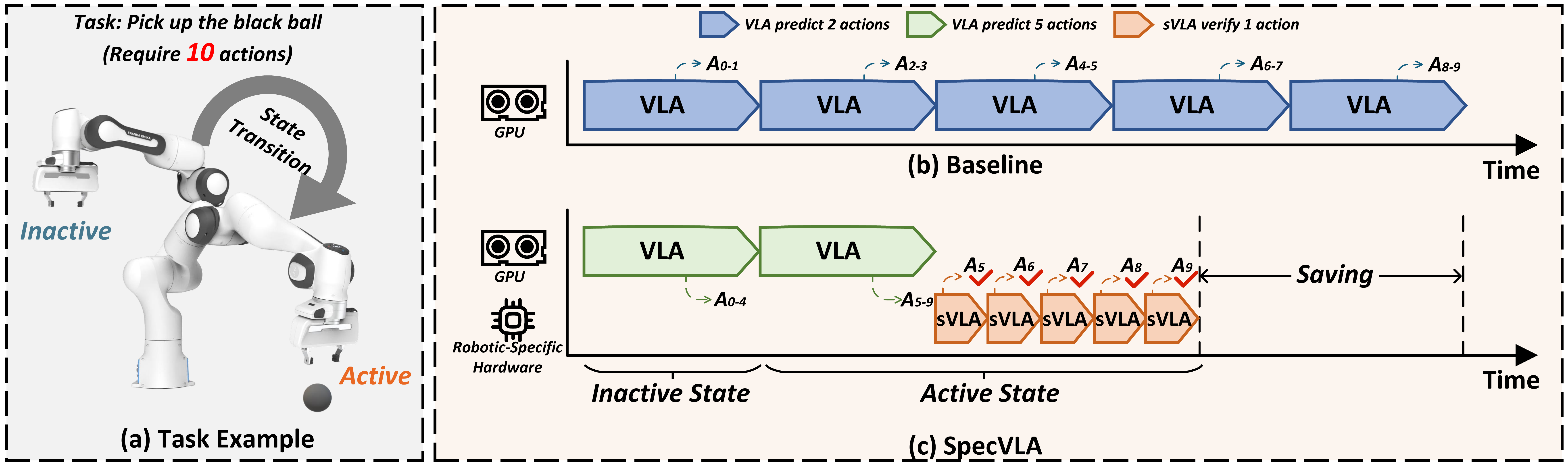}
\vspace{-0.3cm}
\caption{Overview of SpecVLA: (a) A task example of a robot; (b) Existing VLA algorithm; (c) SpecVLA paradigm.}
\vspace{-0.3cm}
\label{fig-spvla_intro}
\end{figure*}
\noindent\textbf{Our Approach:} To achieve the best of both worlds: amortized inference latency through long sequences and high task success rate, we present SpecVLA, an algorithm–system co-design framework for real-time action predictions with adaptive sequence length.
% that combines the predictive strength of VLA with the efficiency of sVLA through a speculative–verify execution paradigm.
In inactive states, SpecVLA invokes the full VLA to speculatively predict long action sequences (e.g., 5 actions per inference in the example of Fig.~\ref{fig-spvla_intro}), amortizing the inference cost. In active states, SpecVLA leverages the compact sVLA to verify each action before execution. If verification fails, the system immediately rolls back and re-invokes VLA to regenerate the action sequence. As manifested in Fig.~\ref{fig-spvla_intro}, this collaborative mechanism allows SpecVLA to issue long-action-length predictions with lightweight safety checks, achieving the efficiency of speculative execution while maintaining the correctness of reactive control.

\noindent\textbf{System Design: \emph{Speculative Dataflow with Robotic-Specific Hardware.}} To translate the theoretical performance improvement of SpecVLA into real robotic system efficiency, we design a heterogeneous system consisting of a GPU and a robotic-specific accelerator. The GPU is for full-precision VLA inference, and the dedicated robotic hardware is for sVLA. In particular, as illustrated in Figure~\ref{fig-spvla_intro}, when the robot operates in the active state, the sVLA inference is performed at high frequencies for short-sequence verification, while the VLA inference is invoked at lower frequencies for long action prediction. To hide the latency of frequent sVLA calls, VLA and sVLA are executed on separate hardware as aforementioned. We introduce a \emph{speculative dataflow} that decouples the algorithmic dependencies between VLA and sVLA. The key idea is to allow VLA inference to proceed \emph{speculatively}, without waiting for the verification result produced by sVLA. 
The technical contributions of this paper are as follows:

\begin{itemize}
\item A algorithm-system co-designed framework for adaptive robotic action prediction by taking advantage of the continuity of the environmental frames.
\item SpecVLA proposes a runtime algorithm to systematically identify the active and inactive states so that the predicted action sequence length, as well as its verification, can be determined as needed.
%[numbers here, how accurate how effective this is?]
%differential residuals
\item SpecVLA proposes a hardware-friendly VLA quantization method based on differential computation and block-wise mixed precision quantization. %[numbers here, how much accuracy degradation and how much compute savings?]
\item A novel dataflow is introduced to coordinate across the full-precision and the quantized model to hide the latency of each other, while verifying the correctness of the predicted actions effectively. %[numbers here, how much performance improvement by hiding the latency? how much data traffic saved?]
\end{itemize}

Our evaluation shows that SpecVLA achieves a $2.9\times$ and $1.9\times$ end-to-end speedup over NVIDIA A100 GPU and the state-of-the-art VLA acceleration framework Dadu-Corki, while maintaining a comparable success rate.

\section{Background and Motivation}\label{sect:background}

\subsection{Basics of VLA Model}

The Vision-Language-Action (VLA) model has shown remarkable potential in embodied intelligence, demonstrating exceptional performance even in complex environments. This capability positions VLA as a promising research avenue toward achieving Artificial General Intelligence (AGI). In this section, we take OpenVLA~\cite{kim2024openvla}, a state-of-the-art VLA model, as a representative example to illustrate its basic structure and execution process.

\begin{figure}[h]
\centering
\includegraphics[width=0.9\linewidth]{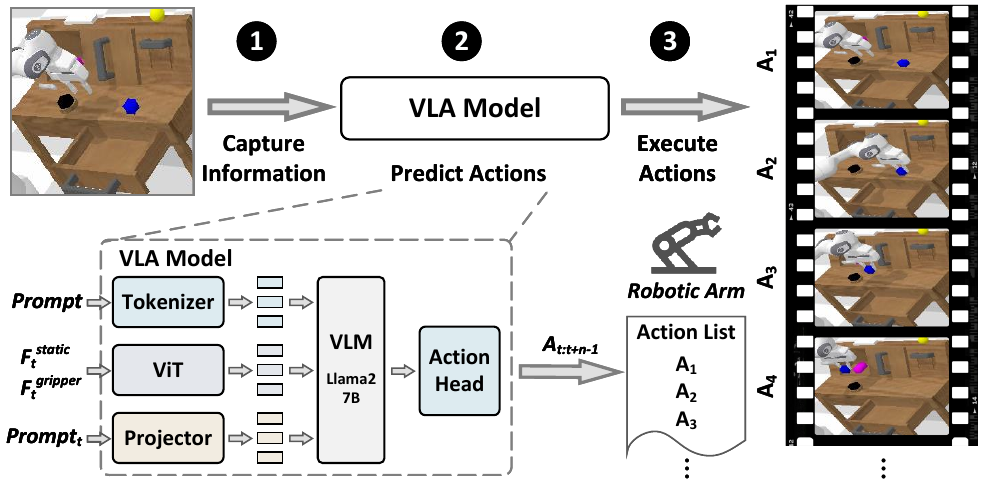}
% \vspace{-0.4cm}
\vspace{-0.2cm}
\caption{The execution process of the VLA model.}
\vspace{-0.4cm}
\label{fig-background-vla}
\end{figure}

The VLA model is primarily designed to control a 6- or 7-degree-of-freedom (DoF) robotic arm equipped with a gripper. Given a fixed action length of $n$, the execution process of the VLA model, illustrated in Fig.~\ref{fig-background-vla}, consists of three stages. In the first stage, the sensors equipped in the robot arm capture the current environmental information, including a global image, a local image, and the robot's current state (e.g., joint angles, gripper coordinates, etc.). Next, the VLA model predicts $n$ future actions based on the acquired sensory data and a language prompt (denoted as $Prompt$). Finally, the gripper sequentially executes the $n$ predicted actions. This execution process is repeated iteratively until the task is successfully completed.

Fig.~\ref{fig-background-vla} also shows the structure of the VLA model, which contains five modules: (1) a Vision Transformer (ViT) model, which is responsible for mapping the captured image to visual tokens; (2) a Projector, which consists of two linear layers and a GELU activation function to map the robot state to state tokens; (3) a tokenizer, designed to convert $Prompt$ to text tokens; (4) a VLM Backbone (such as Llama 2 7B), which packages visual tokens, state tokens and text tokens as input and extracts feature information from them; (5) an action head, which consists of multiple MLP blocks and decodes the feature representations from the VLM backbone to generate the predicted action sequence $A_{t:t+n-1}$, where $t$ is the current timestep.

\subsection{VLA Accelerator Design}

To enhance the efficiency of embodied AI systems, Dadu-Corki~\cite{DBLP:conf/isca/HuangH0Y0M00L0G25} proposes a novel execution pipeline aimed at reducing end-to-end latency. It offloads VLA inference to a cloud server, while leaving robot action execution to the robot itself. This allows VLA inference to leverage the computing power of a GPU server. Additionally, Dadu-Corki accelerates robot action execution by using a dedicated accelerator that quickly converts trajectories into control signals. The accelerator also maximizes intermediate data reuse through customized circuits and data pipelines. Although this execution flow introduces the overhead of transmitting newly captured frames to the server, Dadu-Corki streamlines this transmission by sending frames concurrently with the robot's execution process. By combining these techniques, Dadu-Corki effectively hides communication latency and reduces VLA inference latency, leading to a significant reduction in end-to-end latency. Consequently, we build our proposed SpecVLA upon Dadu-Corki, where both VLA and sVLA inference are performed on the cloud server.

\noindent\textbf{Limitations of Dadu-Corki:} While Dadu-Corki reduces the latency of VLA inference and robot control through a dedicated execution pipeline and specialized circuits, it fails to leverage the interaction between the robot and its environment, which results in a relatively short predicted action length ($< 4$ actions per inference). In contrast, recognizing that the prediction capability of VLAs is closely tied to the current environment, we propose the Speculative-Verification execution (SpecVLA) paradigm. SpecVLA focuses on increasing the number of actions per VLA inference ($8$ actions per inference) in simpler environments, while utilizing an efficient VLA variant, sVLA, for action verification when the robot enters a more complex environment. By doing so, SpecVLA effectively balances action length and task success rate.

\subsection{Motivation}\label{sect:motivation}

To enable efficient robotic manipulation, we conduct a comprehensive analysis of representative VLAs, including OpenVLA~\cite{kim2024openvla} and RDT~\cite{liu2024rdt}. Based on this analysis, we identify the performance bottlenecks and key observations.

\begin{figure}[h]
\centering
\includegraphics[width=\linewidth]{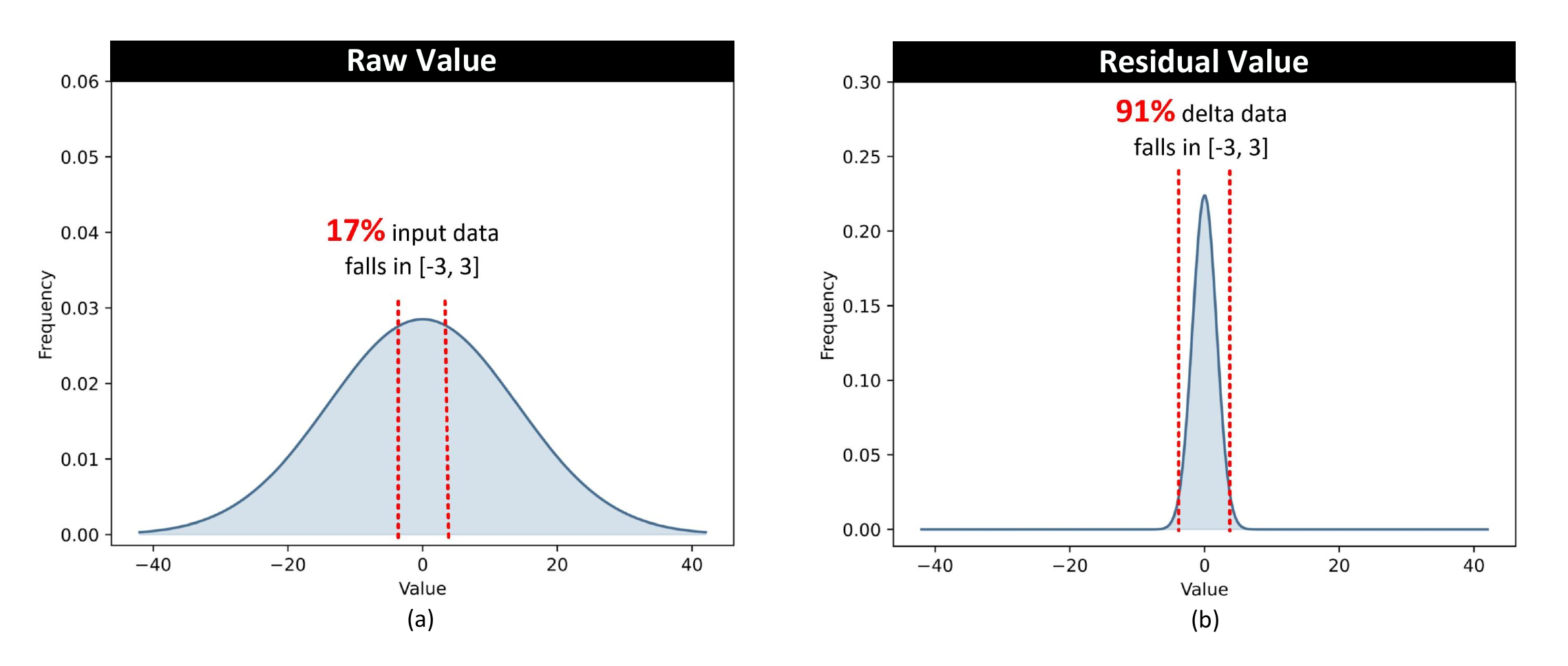}
\vspace{-0.8cm}
\caption{Data distribution of (a) raw features, and (b) residual features.}
\vspace{-0.2cm}
\label{fig-background-distribute}
\end{figure}

\textbf{Challenge 1: \textit{High Computational Cost Makes VLAs Too Slow for Real-Time Control.}} Full-precision VLA models require more than 3 TFLOPs, resulting in per-inference latencies far exceeding real-time requirements. Reducing computation via quantization is a natural direction, but naively applying quantization often destroys accuracy.

\textbf{Observation 1: \textit{Strong Temporal Similarity Enables a Lightweight sVLA.}}
Due to the limited displacement of the gripper between consecutive frames, the inputs to successive VLA inferences exhibit high temporal redundancy. After subtracting frame-to-frame features, the residuals have a significantly narrower distribution (as shown in Fig.~\ref{fig-background-distribute}), making them well suited for mixed-precision differential-based quantization. This insight motivates constructing a lightweight VLA model (sVLA) using block-wise residual quantization. Although this reduces inference latency substantially, sVLA alone cannot reliably predict long action sequences.

\begin{figure}[h]
\centering
\includegraphics[width=0.9\linewidth]{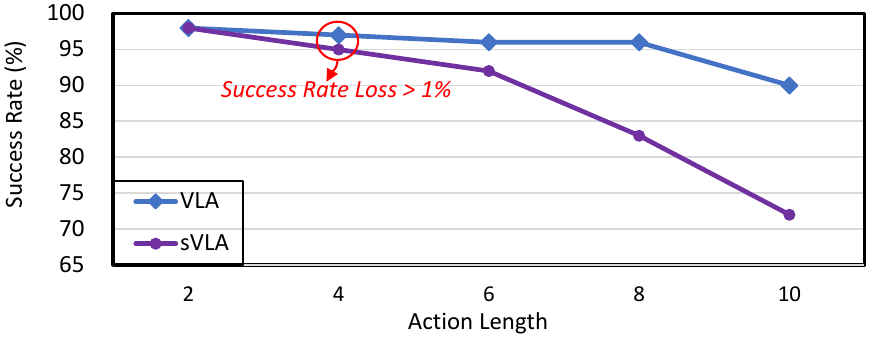}
\vspace{-0.4cm}
\caption{Success rate of VLA and sVLA under different action length.}
\vspace{-0.2cm}
\label{fig-background-actionlength}
\end{figure}

\textbf{Challenge 2: \textit{sVLA Suffers From Accuracy Degradation at Longer Action Prediction Length.}} 
We evaluate the success rate of sVLA under varying action lengths in Fig.~\ref{fig-background-actionlength}. The success rate is defined as the ratio of tasks that are successfully completed to the total number of tasks, and the action length denotes the number of low-level actions predicted by a single inference call. Compared with the original VLA, the success rate remains comparable when the predicted action length is fewer than two actions but drops by an average of $9\%$ as the sequence length grows. Because the end-to-end latency of a robotic system depends on both inference speed and the number of predicted actions per inference, such limited action length still prevents timely and responsive robot control.

\begin{figure}[h]
\centering
\includegraphics[width=0.9\linewidth]{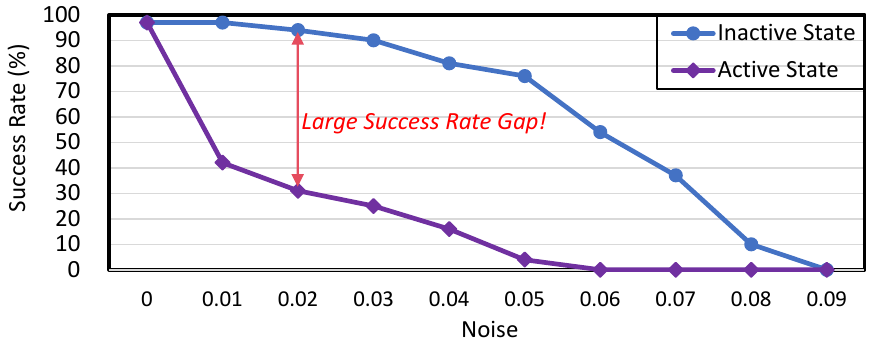}
\vspace{-0.4cm}
\caption{Explore environmental state: success rate under different noise patterns.}
\vspace{-0.3cm}
\label{fig-background-noise}
\end{figure}

\textbf{Observation 2: \textit{Prediction Capability of VLAs Strongly Correlates with Environmental State.}} By analyzing the motion trajectories of the robot's gripper, we categorize the environmental state into two levels: the \textbf{active state} and the \textbf{inactive state}. The active state corresponds to periods when the gripper interacts with or approaches the target object, during which the object's position or orientation may change, critically affecting task success. In contrast, the inactive state refers to gripper movements between objects without direct interaction, where actions exert minimal influence on the task outcome. To validate this distinction, we introduce different noise levels into the actions of both states and evaluate the OpenVLA model on the LIBERO dataset. As shown in Fig.~\ref{fig-background-noise}, the success rate in the inactive state (blue curve) consistently exceeds that in the active state (purple curve), confirming that the active state is more sensitive to perturbations. Accordingly, the system should identify the active state and employ sVLA to verify the reliability of each predicted action before physical execution. During the inactive state, the robot can instead use VLA to predict longer action sequences and execute them directly without verification, thereby improving overall efficiency.

% we introduce different noise levels into the actions of both states and evaluate the OpenVLA model on the LIBERO dataset.

\begin{figure}[h]
\centering
\includegraphics[width=0.9\linewidth]{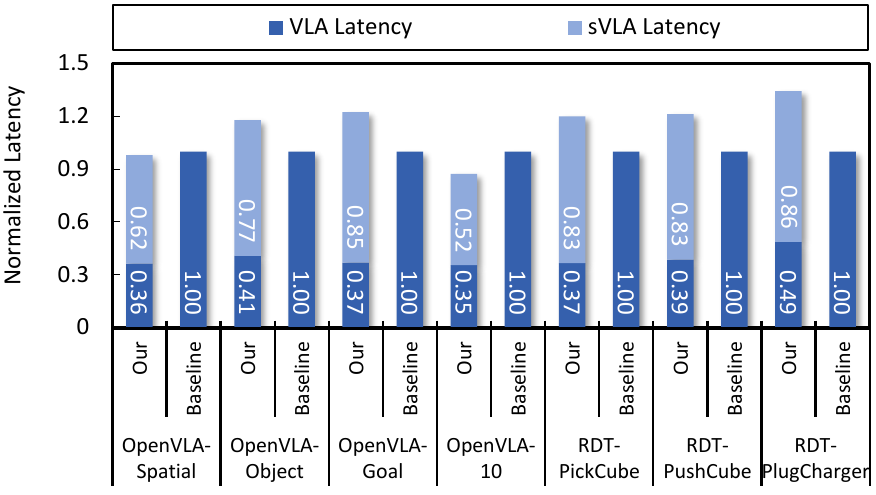}
\vspace{-0.2cm}
\caption{The latency of SpecVLA and baseline VLA model on A100 GPU.}
\vspace{-0.2cm}
\label{fig-background-svla-vla}
\end{figure}

\textbf{Challenge 3: \textit{Sequential VLA–sVLA Execution Causes Poor Hardware Utilization.}} When the robot operates in the active state, sVLA is always executed after VLA for action verification, introducing a strong data dependency between the two models. Meanwhile, since sVLA is invoked more frequently than VLA, this sequential execution pattern inevitably increases the overall latency. To quantify this effect, we implement sVLA and measure the runtime of both VLA and sVLA on an NVIDIA A100 GPU. As shown in Fig.~\ref{fig-background-svla-vla}, the combined latency of VLA and sVLA exceeds that of the baseline VLA model alone. This is primarily caused by the serialized execution flow between the two models, which prevents the theoretical performance gains from translating into practical acceleration.

%We evaluate the inference latency of both VLA and sVLA models on the A100 GPU. As shown in Fig.~\ref{}, the sVLA model achieves $2\times$ speedup over the VLA model, far away from the expected latency reduction ($4\times$), given that we use 4, incurring non-negligible overhead. This discrepancy arises because the GPU does not efficiently support the proposed mixed-precision residual computation, preventing the translation of theoretical gains into practical acceleration.

% This discrepancy arises because the GPU does not efficiently support the proposed mixed-precision residual computation, preventing the translation of theoretical gains into practical acceleration.
% The root cause of this problem is that sVLA and VLA are executed serially

\textbf{Observation 3: \textit{Cloud Platforms Provide Natural Heterogeneity for Parallelization.}}
Modern cloud servers typically integrate GPUs and NPUs, enabling the potential to parallelize VLA and sVLA execution across heterogeneous devices. Motivated by this, we design a heterogeneous computing system in which the GPU performs VLA inference while the NPU handles sVLA inference. To exploit the hardware parallelism, we propose a speculative dataflow that decouples the inherent data dependency between VLA and sVLA, enabling simultaneous execution on both devices. 
\section{SpecVLA Algorithm}\label{sect:ctq-algo}
In this section, we will propose the  Speculative VLA Inference and Action Verification (SpecVLA) algorithm and the sVLA construction method to reduce the end-to-end latency of the embodied AI system.

\subsection{Overview of SpecVLA}

The objective of the SpecVLA framework is to generate action sequences of maximal possible length while maintaining an acceptable task success rate. To achieve this, SpecVLA first predicts the current environmental state of the robot and then adaptively adjusts the action length by coordinating the execution of VLA and sVLA:

1. \textbf{Active State.} In the active state, where upcoming actions are expected to have a significant impact on task success, SpecVLA first conducts the VLA inference to predict a long sequence of actions, referred to as Predicted Actions, with a predefined length $N$. The parameter $N$ serves as a hyperparameter that jointly affects both inference efficiency and task success rate, and its optimal setting is explored in Section~\ref{sect:experiment}. Before each Predicted Action is executed, SpecVLA triggers the sVLA model to generate an Expected Action based on the latest sensor image. The Predicted Action is then compared with the Expected Action to assess its reliability. If the two actions are sufficiently similar, the Predicted Action is accepted and executed by the robot arm. Otherwise, the prediction is considered unreliable, and the VLA model is re-invoked to generate a new action sequence.

2. \textbf{Inactive State.} In contrast, during the inactive state, where actions have limited influence on task outcome, SpecVLA relies solely on VLA to produce a long sequence of Predicted Actions without performing action verification, thereby improving inference efficiency.

By combining environmental state prediction, speculative VLA inference, and sVLA action verification, SpecVLA adaptively maximizes the effective prediction length while maintaining a high task success rate.

\begin{figure*}[h]
\centering
\includegraphics[width=\linewidth]{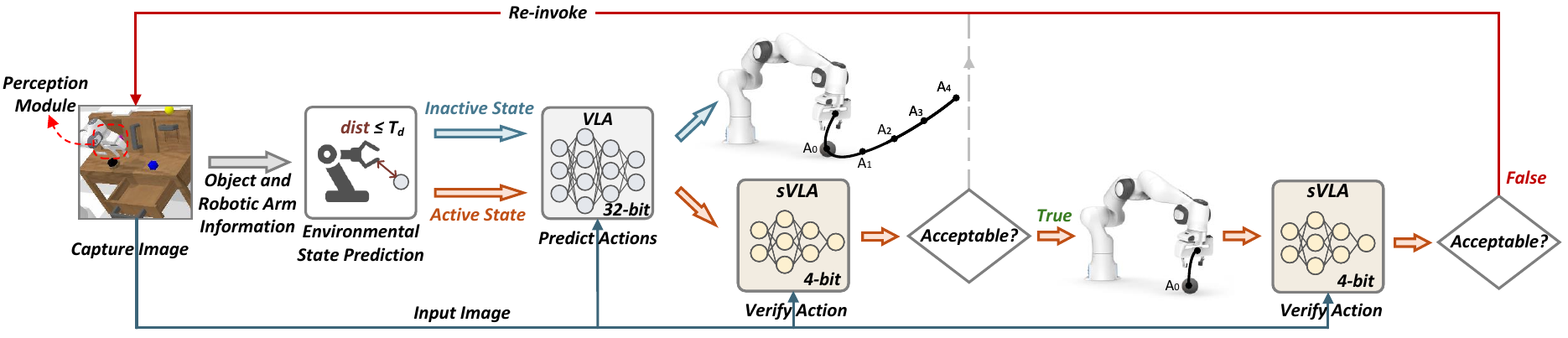}
\vspace{-0.7cm}
\caption{The execution process of SpecVLA.}
\vspace{-0.4cm}
\label{fig-algorithm-speculative-prefilling}
\end{figure*}

The execution process of SpecVLA is detailed in Fig.~\ref{fig-algorithm-speculative-prefilling}, where $A_i$ denotes the $i$-th Predicted Action. Initially, SpecVLA captures the environmental state and then invokes VLA to generate a prediction sequence consisting of five actions, denoted as $A_{0:4}$. If the current environment is classified as inactive (see the top part of the figure), the robot controller directly treats $A_{0:4}$ as execution commands and executes them without additional verification. If the environment is classified as active (the bottom part of the figure), SpecVLA activates the sVLA model to generate the Expected Action and compares it with $A_0$ to assess its reliability. Since $A_0$ closely matches the Expected Action, the Predicted Action is deemed acceptable and executed by the robot arm. This verification continues until $A_4$ significantly deviates from the Expected Action, prompting SpecVLA to re-invoke the VLA model to generate a new action sequence, $A_{4:8}$. Meanwhile, the system re-evaluates the environmental state.

%The detailed execution process of SpecVLA is illustrated in Fig.~\ref{fig-algorithm-speculative-prefilling}, where $A_i$ represents the Predicted Action at time $T_i$, and $A_i^{'}$ denotes the corresponding Expected Action. As shown in the figure, at time $T_i$, SpecVLA first captures environmental information to identify the current state. It then invokes the VLA model to generate a prediction sequence consisting of eight actions, denoted as $A_{i:i+7}$. Since the environment at $T_i$ is classified as active, SpecVLA activates the sVLA model to produce the Expected Action $A_i^{'}$ based on the captured image at that moment. The Predicted Action $A_i$ is then compared with $A_i^{'}$ to assess its reliability. Because $A_i \approx A_i^{'}$, the Predicted Action is considered acceptable and executed by the robot arm. This verification process continues until time $T_{i+5}$, where $A_{i+5} \neq A_{i+5}^{'}$, prompting SpecVLA to re-invoke the VLA model to generate a new action sequence $A_{i+5:i+12}$. Meanwhile, the system re-evaluates the environment and determines that it has transitioned to an inactive state. Consequently, the newly predicted actions $A_{i+5:i+12}$ are directly fed into the robot controller and serve as execution commands without additional verification.

\subsection{Technical Details of SpecVLA}

To fully realize the potential of SpecVLA, two key challenges must be addressed. The first challenge involves determining the current environmental state to adaptively adjust the length of Predicted Actions. The second challenge is how to accurately assess whether a Predicted Action is sufficiently precise and safe for execution.

\subsubsection{Environmental State Prediction}

To address the first challenge, we predict the environmental state by analyzing both the distance between the robot arm and surrounding objects and the current state of the robot arm. The detailed procedure is outlined in Algorithm~\ref{alg:task_state_identification}. First, the algorithm computes the Euclidean distance $D_i=\sum_{k=1}^{3} (G_c[k]-O_c^{i}[k])^2$ between the robot arm and each object in the environment (Lines 2–3), where $G_c$ and $O_c$ are the three-dimensional coordinates of the robot arm and each object, respectively, as provided by the robotic system’s onboard perception sensors. These distances are then compared against a predefined threshold $T_d$ to assess whether any objects are too close to the robot arm (Lines 4–5). Moreover, the environmental state is also related to the robot arm state $G_s$. If the robot arm is open and no nearby objects are detected, it suggests that the robot arm is neither currently interacting with an object nor about to do so. Hence, the task is classified as being in the inactive state (Lines 6–7). Conversely, if the robot arm is close to at least one object, it implies that the robot arm is either holding an object or is very likely to grasp one soon, and the task is categorized as being in the active state (Lines 8–9).
%The distance threshold Td is set to a conservatively large value to reduce unexpected interaction with the arm body or tools.
%We will discuss how to select an optimal $T_d$ in the experiment section.
%\textcolor{brown}{We set the threshold $T_d$ to a conservatively large value to minimize unexpected interactions with the robot arm or tools. The methodology for selecting the optimal $T_d$ will be further discussed in the experiment section.}

\begin{algorithm}[!t]
    % \setstretch{0.975}
    % \footnotesize
    \caption{Environmental State Prediction.}
    \label{alg:task_state_identification}
    \KwIn{Robot arm state $G_s$\;
    \qquad \quad Object coordinate set $\{O_c\}$\; 
    \qquad \quad Robot arm coordinate $G_c$\; 
    \qquad \quad Distance threshold $T_d$\;}
    \KwOut{Environmental state $S_{env}$.}
    % $T_s\leftarrow 1$\;
    $n \leftarrow 0$\;
    \For{$i\leftarrow 0$ to $|\{O_c\}|$}{
        $D_i\leftarrow\texttt{$Distance$}(O_c^i,G_c)$\;
        \uIf{$D_i \leq T_d$}{
            $n\leftarrow n + 1$\;
        }
    }
    \If{$G_s == -1$ and $n < 1$}{
        $S_{env} \leftarrow Inactive State$\;
    }    
    \Else{
        $S_{env} \leftarrow Active State$\;
    }
    \Return{$S_{env}$}\;
    % \Return{$M$}\;
\end{algorithm}

\subsubsection{Action Verification}

To address the second challenge, we begin by performing sVLA inference to get an Expected Action. Subsequently, we calculate the L1 distance between the Predicted Action $A$ and the Expected Action $A^{'}$. A lower distance value indicates that the two actions are functionally close, while a higher value suggests significant discrepancies, implying that the Predicted Action may be inaccurate. Specifically, we apply min–max normalization to each action prior to computing the L1 distance to ensure comparability across heterogeneous action space. Each action consists of seven dimensions, corresponding to three-dimensional positional changes, three-dimensional rotational changes, and a one-dimensional gripper state. For each action dimension $A[k]$, we record its maximum and minimum values ($A\_high[k]$ and $A\_low[k]$) according to the robot's control interface. The action is then normalized as: 
\[
\tilde{A}[k] = \frac{A[k] - A\_low[k]}{A\_high[k] - A\_low[k]}
\]
Considering that different action dimensions exhibit varying degrees of sensitivity during complex manipulation, weighted L1 distance is then computed over the scaled action to modulate the verification sensitivity. By calculating the standard deviation $\sigma[k]$ for each normalized dimension, we assign a weight $w[k] = \frac{1}{\sigma[k]}$ to balance their relative contributions: 
\[
L_1(\tilde{A}, \tilde{A}') = \sum_{k=1}^{7} w[k]\times|\tilde{A}[k] - \tilde{A}'[k]|,\ w[k]=\frac{1}{\sigma[k]}
\]
This weighting strategy ensures that dimensions with high intrinsic variance receive lower weights as they exhibit higher error tolerance, while priority is given to highly stable dimensions given their criticality for high-precision tasks. Such a scheme significantly enhances the accuracy and reliability of the action validation.

Finally, the computed distance is compared against a predefined threshold $T_s$: if the distance is less than $T_s$, the two actions are considered similar and the Predicted Action is accepted; otherwise, they are regarded as dissimilar, and the Predicted Action is rejected. $T_s$ will be explored in the experiment section.

%Specifically, we conduct $L_1(Action,Action^{'})=\sum_{k=1}^{7} |Action[k]-Action^{'}[k]|$, where $Action[k]$ is one of the seven parameters within each action.
%This weighting strategy ensures that dimensions with high intrinsic variance (often containing more noise) receive lower weights, while highly stable dimensions—which are typically critical for high-precision tasks—are prioritized.

\subsection{Construction of sVLA}

\begin{figure}[h]
\centering
\includegraphics[width=\linewidth]{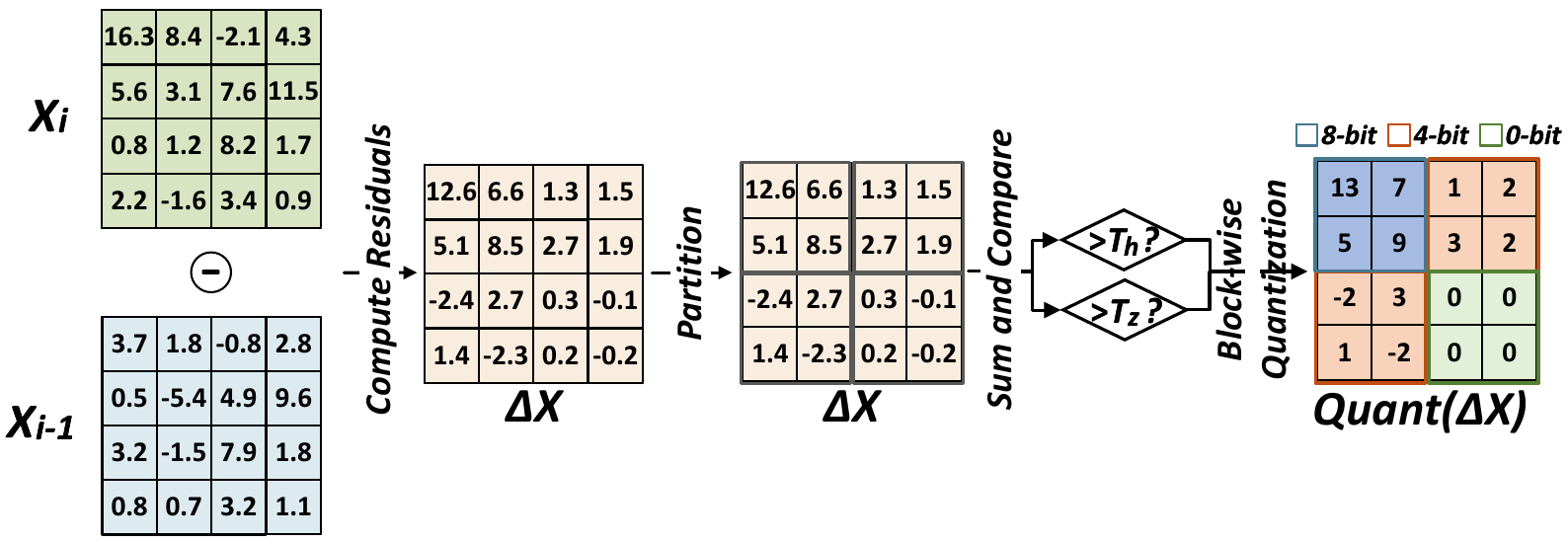}
\vspace{-0.6cm}
\caption{The process of the differential-based, hardware-friendly quantization.}
\vspace{-0.2cm}
\label{fig-algorithm-svla}
\end{figure}

To enable efficient execution of the SpecVLA method, we construct sVLA using a differential-based, hardware-friendly quantization scheme, as illustrated in Fig.~\ref{fig-algorithm-svla}. At each time step $T_i$, we first compute the residual feature $\Delta X_i$ by subtracting the current input feature $X_i$ from the previous input feature $X_{i-1}$: $\Delta X_i = X_i - X_{i-1}$. A straightforward approach to processing these residuals is to apply mixed-precision quantization according to their magnitudes. However, this strategy introduces complex hardware control and leads to poor resource utilization when handling high- and low-precision residuals simultaneously.

\begin{figure*}[!t]
\centering
\includegraphics[width=0.9\linewidth]{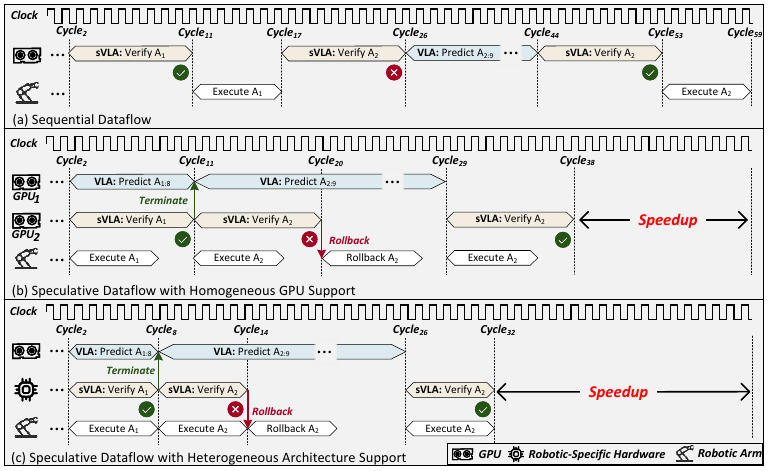}
\vspace{-0.4cm}
\caption{Details of SpecVLA dataflow: (a) Sequential dataflow; (b) Speculative dataflow with homogeneous gpu support; (c) Speculative dataflow with heterogeneous architecture support.}
\vspace{-0.4cm}
\label{fig-system-dataflow}
\end{figure*}

To overcome this issue, we propose a block-wise quantization scheme, which partitions the residual matrix into fixed-size blocks. For each residual block, we compute the sum of the absolute values for all elements and compare it with two thresholds, $T_z$ and $T_h$ ($T_z < T_h$), to determine its quantization bitwidth: 1.if the sum is less than $T_z$, the block is considered insignificant and all values are set to zero; 2.if the sum exceeds $T_h$, the block is regarded as important and quantized in high precision; 3.otherwise, the block is assigned low-precision quantization.
%we compute the sum of all elements and compare it with two thresholds

After block-wise quantization, the residual matrix is compressed by removing zero blocks. The high- and low-precision residual blocks are then multiplied with the corresponding weights to obtain the residual output. The final output is computed by accumulating this residual output with that from the previous time step $T_{i-1}$. This quantization-and-accumulation process is applied layer-wise across sVLA, enabling fast and accurate inference with high hardware efficiency.

\section{SpecVLA Heterogeneous Architecture}

In this section, we will introduce the SpecVLA heterogeneous architecture to efficiently support the SpecVLA algorithm.

\subsection{Speculative Dataflow}

We use Fig.~\ref{fig-system-dataflow}(a) to illustrate the sequential dataflow, where the SpecVLA algorithm is executed on a cloud GPU while the robot performs actions locally. In this sequential design, all stages—including VLA prediction, sVLA verification, and action execution—exhibit strict data dependencies and are therefore serialized. This is because prediction depends on newly observed images after execution, verification depends on predicted actions, and execution depends on verification results. This fully sequential dataflow leads to high end-to-end latency and poor hardware utilization.

%, resulting in high latency and poor hardware utilization. In principle, prediction, verification, and execution have strict data dependencies: each VLA inference conditions on newly observed images after the previous action execution.

\noindent\textbf{Necessity of a Speculative Dataflow.} These limitations highlight the need for a speculative dataflow, which decouples the dependency between VLA and sVLA by executing VLA inference \textbf{\emph{speculatively}}, in advance of sVLA verification. To implement this design, a straightforward solution is to use two GPUs connected via NVLink for VLA prediction and sVLA verification, as shown in Fig.~\ref{fig-system-dataflow}(b). GPU$_2$ performs sVLA inference to generate the Expected Action, while GPU$_1$ concurrently runs VLA inference to speculatively predict the next action sequence. Once GPU$_2$ completes sVLA inference, it compares the prior Predicted Action with the newly generated Expected Action. If the prior Predicted Action is inaccurate—indicating that VLA inference is necessary—GPU$_2$ signals GPU$_1$ to continue and complete the remaining VLA computation. Otherwise, GPU$_2$ terminates the speculative VLA execution on GPU$_1$ early. As shown in Fig.~\ref{fig-system-dataflow}(b), at cycle $2$, GPU$_2$ executes sVLA to obtain the Expected Action $A_1'$, while GPU$_1$ speculatively computes the predicted action sequence $A_{1:8}$. At cycle $11$, after verification confirms that $A_1$ is accurate, GPU$_2$ terminates the speculative VLA execution on GPU$_1$ and allows GPU$_1$ to proceed to the next speculative round, predicting a new action sequence $A_{2:9}$. At cycle $20$, since $A_2$ is inaccurate, GPU$_2$ signals GPU$_1$ to continue the remaining VLA computation until cycle $29$. In this example, GPU$_1$ saves the latency of VLA computation from cycle $11$ to cycle $20$ by overlapping it with GPU$_2$'s sVLA execution. Furthermore, since the robot arm's movement latency is non-negligible, we further parallelize the robot arm's movement with the computations performed by GPUs. To ensure correct physical state progression when sVLA verification rejects a predicted action, we incorporate a rollback mechanism. If verification fails, SpecVLA issues compensatory reverse motions to restore the robotic arm to its last committed state. Physically irreversible effects (e.g., full gripper closure) require multi-step execution, whereas SpecVLA performs verification after at most one motion primitive. This granularity remains well below the threshold for triggering irreversible state transitions, ensuring safe recovery.

\begin{figure}[h]
\centering
\includegraphics[width=0.9\linewidth]{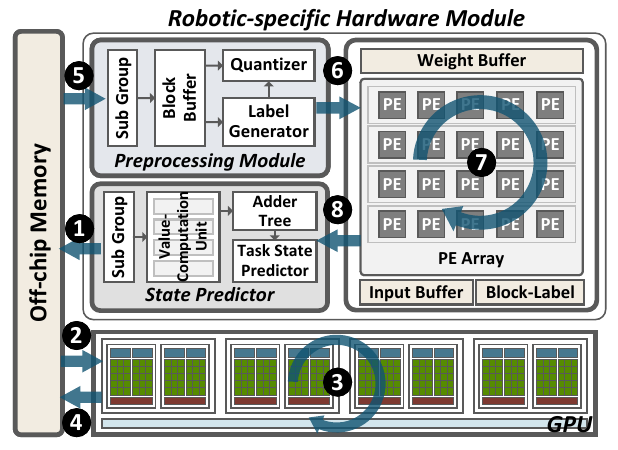}
\vspace{-0.4cm}
\caption{The overview of the SpecVLA heterogeneous architecture.}
\vspace{-0.4cm}
\label{fig-system-overview}
\end{figure}
%\textcolor{red}{As a result, the speculative dataflow hides sVLA inference behind VLA inference, reducing the overall latency by approximately 29\%.}

%. This inefficiency stems from two key factors. First, the SIMT architecture of GPUs is fundamentally ill-suited for the control-intensive operations required by the SpecVLA algorithm. Second,
However, the speculative dataflow using only GPUs struggles to achieve peak performance. This inefficiency stems from two key factors. First, the block-wise quantization used in sVLA is difficult to map efficiently onto modern GPUs. The frequent precision switching across blocks triggers repeated kernel launches, incurring an additional overhead. Second, the SIMT architecture of GPUs is fundamentally ill-suited for the control-intensive operations required by the SpecVLA algorithm, such as environmental state prediction. These limitations motivate the need for a dedicated robotic-specific hardware module tailored to the SpecVLA algorithm, with specialized support for fast sVLA verification and environmental state prediction. With this hardware support, the overall latency is further reduced, as illustrated in Fig.~\ref{fig-system-dataflow}(c).

\subsection{Robotic-Specific Hardware Design}

% \begin{figure}[h]
% \centering
% \includegraphics[width=0.9\linewidth]{figures/specvla_arch_overview.pdf}
% \vspace{-0.3cm}
% \caption{The overview of the SpecVLA heterogeneous architecture.}
% \label{fig-system-overview}
% \end{figure}

Fig.~\ref{fig-system-overview} illustrates the SpecVLA heterogeneous architecture, which consists of a GPU and a robotic-specific hardware module. To handle sVLA inference, the robotic-specific hardware module applies a block-splitting strategy to unify the precision within each residual block and leverages a SIMD-based PE array to accelerate block-level matrix multiplications. In addition, the module integrates two key components: a state predictor and a preprocessing module. The state predictor identifies the current environmental state and computes the L1 distance between the Predicted Action and the Expected Action in a reconfigurable manner. And the preprocessing unit is responsible for generating residuals and performing block-level quantization.

The dataflow of the SpecVLA heterogeneous architecture operates as follows. The state predictor first realizes the environmental state prediction operator and identifies the current state (Step \Large{\ding{182}}\normalsize{ }). Following this, the GPU loads the input and weight data of the VLA model from DRAM (Step \Large{\ding{183}}\normalsize{ }), performs VLA inference (Step \Large{\ding{184}}\normalsize{ }), and writes the resulting Predicted Actions back to DRAM (Step \Large{\ding{185}}\normalsize{ }). In the meantime, the robotic-specific hardware module runs the sVLA inference (Step \Large{\ding{186}}\normalsize{ }-\Large{\ding{188}}\normalsize{ }) to generate the corresponding Expected Action. Once both Predicted Action and Expected Action are available, the state predictor realizes the action verification operator and compares the Predicted Action with the Expected Action to verify its correctness (Step \Large{\ding{189}}\normalsize{ }).
% If the prediction is correct, the robotic-specific hardware module verifies the next Predicted Action. Otherwise, it instructs the robotic arm to rollback the inaccurate Predicted Action and waits for the GPU to complete the new VLA prediction.

\subsubsection{SIMD-based PE Array}

To simplify the hardware control required for handling multiple precisions, the preprocessing module (details in Section~\ref{sect:preprocess}) decomposes each high-precision residual block into a high-bit block and a low-bit block, while zero blocks are removed via feature condensation. After decomposition and condensation, all components share the same bitwidth as native low-precision blocks, allowing the SIMD-based PE array to treat all blocks uniformly and operate using a single low-precision datapath. To support computation under this block-splitting scheme, the PE array is optimized for low-precision block-level operations. Each PE unit inside the array contains $G$ low-precision multipliers, a $\log_2 G$-level adder tree, and an accumulator register, collectively implementing a vectorized inner product. A block-label is also generated by the preprocessing module and attached to each block, which indicates whether the block is a native low-precision block, the low-bit component of a high-precision block, or the high-bit component of a high-precision block. Based on this label, the PE unit decides how to accumulate results:
\begin{itemize}
    \item For low-precision blocks and low-bit components, the result is directly accumulated into the register;
    \item For high-bit components, the partial sum is left-shifted before accumulation to ensure numerical correctness.
\end{itemize}

A remaining challenge for the PE array is the substantial on-chip buffer bandwidth required to feed all PEs in parallel. Since every PE would request data simultaneously, the buffer interface can easily become a bottleneck. To mitigate this issue, we group each row of PEs into a PE line, where residual blocks are shared across the PEs through hard-wired shift connections.

\subsubsection{State Predictor}\label{sect:state-prediction}

Our analysis of the SpecVLA algorithm reveals two high-frequency operators—environmental state prediction and action verification—that critically impact end-to-end latency. To accelerate them, we examine their compute patterns and identify a key architectural opportunity: both operators share a common distance-computation data path. Exploiting this insight allows us to accelerate both operators while reducing hardware cost. Specifically, environmental state prediction computes the Euclidean distances between the robot arm and nearby objects, and then infers the environmental state based on these distances and the robot's internal state. Similarly, action verification performs sVLA inference and subsequently evaluates the L1 distance between the Predicted Action and the Expected Action. Although the semantics of the two operators differ, their execution pipelines contain identical stages for distance computation. Motivated by this observation, we design a \emph{reconfigurable state predictor} that unifies these shared operations within a single hardware data path. 

\begin{figure}[h]
\centering
\includegraphics[width=0.92\linewidth]{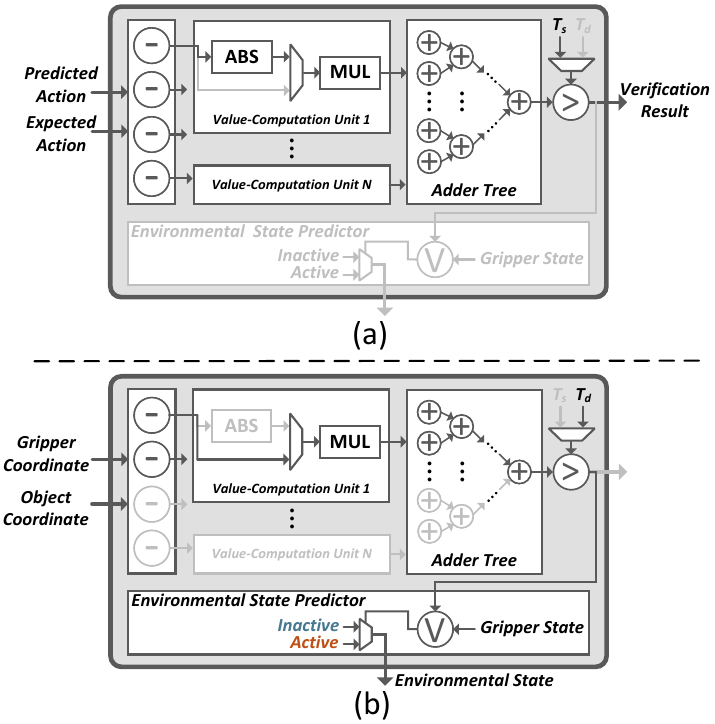}
\vspace{-0.4cm}
\caption{The structure of the state predictor.}
\vspace{-0.7cm}
\label{fig-system-computing-module}
\end{figure}

\textbf{Reconfigurable State Predictor.}
Fig.~\ref{fig-system-computing-module} shows the microarchitecture of the proposed reconfigurable state predictor. The design highlights the core architectural insight: both environmental state prediction and action verification share the same distance-computation pipeline. To exploit this, the predictor integrates a unified datapath composed of a subtractor group, value-computation units, an adder tree, a comparator, an OR gate, and multiple multiplexers. These components are orchestrated through lightweight control signals that dynamically reconfigure the module to support two distinct modes of operation:

\emph{Action Verification Mode.}
As shown in Fig.~\ref{fig-system-computing-module}(a), when the Predicted Action and Expected Action become available, the subtractor group computes their element-wise differences, which are then forwarded to the value-computation units (each applying absolute-then-multiply or squared operations depending on configuration). The adder tree aggregates these values into a single distance metric, which is compared against a threshold $T_s$. If the distance is below the threshold $T_s$, the predictor issues a termination signal to stop the speculative VLA execution on the GPU, ensuring correctness while minimizing wasted computation.

\emph{Environmental State Prediction Mode.}
Fig.~\ref{fig-system-computing-module}(b) illustrates the second configuration. The module reads the gripper coordinates $G_c$ and the $i$-th object coordinates $O_c^i$, computes their Euclidean distance through the same subtract–compute–reduce pipeline, and compares this distance against the threshold $T_d$. If any object resides within the proximity threshold or the gripper state indicates imminent interaction, the predictor classifies the environment as \emph{active}. Otherwise, the environment is classified as \emph{inactive}. Based on the predicted environmental state, the state predictor determines whether to enter the Action Verification Mode to perform action verification in the next round.

Through this dual-mode design, the reconfigurable state predictor unifies two logically distinct operators into a single hardware pipeline, substantially reducing area overhead while accelerating both computations.

\subsubsection{Preprocessing Module}\label{sect:preprocess}

% \begin{figure}[h]
% \vspace{-0.2cm}
% \centering
% \includegraphics[width=0.95\linewidth]{figures/specvla_arch_preprocessing.pdf}
% \vspace{-0.6cm}
% \caption{The structure of the preprocessing module.}
% \vspace{-0.2cm}
% \label{fig-system-preprocessing}
% \end{figure}

The preprocessing module tightly integrates all operators required in the SpecVLA preprocessing pipeline—including residual generation, block-wise quantization, and block-label creation—within a single hardware unit. By co-locating these operators on a unified datapath, the module eliminates unnecessary data movement between separate hardware modules, thereby significantly reducing latency and improving energy efficiency.

% \begin{figure}[h]
% \vspace{-0.2cm}
% \centering
% \includegraphics[width=0.95\linewidth]{figures/specvla_arch_preprocessing.pdf}
% \vspace{-0.6cm}
% \caption{The structure of the preprocessing module.}
% \vspace{-0.6cm}
% \label{fig-system-preprocessing}
% \end{figure}

\begin{figure}[!t]
% \vspace{-0.2cm}
\centering
\includegraphics[width=0.95\linewidth]{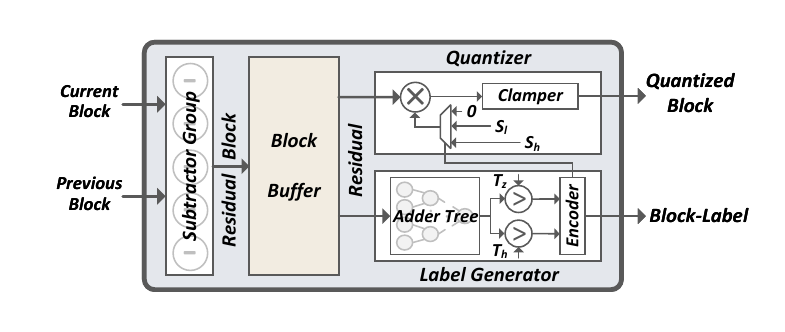}
\vspace{-0.6cm}
\caption{The structure of the preprocessing module.}
\vspace{-0.4cm}
\label{fig-system-preprocessing}
\end{figure}

As illustrated in Fig.~\ref{fig-system-preprocessing}, the module first fetches the current and previous feature blocks from DRAM and forwards them to the subtractor group, which computes the residual blocks. These residuals are streamed into an on-chip block buffer, enabling subsequent operators to process them without any off-chip traffic.
Next, the label generator accumulates all values within each block through an adder tree and determines the block's quantization level by comparing the sum against the thresholds $T_z$ and $T_h$. This comparison yields a block-label that captures the block type: 0 for low-precision block, 1 for low-bit component of a high-precision block, 2 for high-bit component of a high-precision block.
The quantizer then performs the corresponding quantization for each block. For high-precision blocks, it automatically decomposes the block into low-bit and high-bit components, enabling all blocks to be processed using the same low-precision datapath in the PE array.
Finally, the preprocessing module outputs both the quantized blocks and their block-labels directly to the PE array, allowing computation to proceed in a fully pipelined manner without stalling.

% \GarblePages{7}
\section{Evaluation}\label{sect:experiment}
\begin{table*}[h]
\centering

% \scriptsize
\resizebox{\textwidth}{!}{%
\begin{tabular}{c|c|ccccccc}
\toprule
\multicolumn{2}{c|}{Benchmark} 
& OpenVLA-Spatial & OpenVLA-Object & OpenVLA-Goal 
& OpenVLA-10 & RDT-PickCube & RDT-PushCube & RDT-PlugCharger \\
\midrule

\multirow{2}{*}{State Proportion (\%)} 
& Active State   & 24.59 & 24.12 & 22.57 & 27.76 & 26.19 & 26.64 & 28.48 \\
& Inactive State & 75.41 & 75.88 & 77.43 & 72.24 & 73.81 & 73.36 & 71.52 \\

\midrule
\multirow{2}{*}{Rollback Frequency (\%)} 
& All Task        & 8.43 & 7.33 & 8.21 & 9.45 & 8.78 & 9.44 & 15.23 \\
& Complex Task & 12.57 & 12.27 & 13.17 & 14.16 & 11.82 & 12.58 & 16.41 \\

\midrule
\multirow{4}{*}{Effective Action Length} 
& Spec-VLA~\cite{wang2025spec} & 2.94 & 2.38 & 2.14 & 2.10 & - & - & - \\
& HeiSD~\cite{zheng2026heisd} & 4.75 & 4.94 & 4.83 & 4.96 & - & - & - \\
& Dadu-Corki-ADAP~\cite{DBLP:conf/isca/HuangH0Y0M00L0G25} & 4.16 & 4.03 & 4.17 & 3.98 & 3.87 & 3.91 & 3.27 \\
& SpecVLA & 7.58 & 7.47 & 7.52 & 7.13 & 7.14 & 7.08 & 6.12 \\

\midrule
\multirow{3}{*}{Path Length} 
& Non-Speculative Baseline & 148 & 174 & 226 & 492 & 508 & 374 & 990 \\
& MoLe-VLA~\cite{zhang2026mole} & 178 & 204 & 248 & 564 & - & - & - \\
& SpecVLA & 153 & 173 & 230 & 507 & 524 & 361 & 1020 \\

\bottomrule
\end{tabular}
}
\caption{Quantitative Analysis of SpecVLA Algorithm across Different Benchmarks and Metrics.}
% \vspace{-0.8cm}
\vspace{-0.8cm}
\label{tab-eva-spec-analy}
\end{table*}

% 2.94	2.38	2.14	2.1 4.75	4.94	4.83	4.96 178	204	248	564
% \begin{table*}[htbp]
% \centering
% \small
% % 稍微加宽行距，达到与前一个学术表一致的舒展视觉效果
% \renewcommand{\arraystretch}{1.3}
% \setlength{\tabcolsep}{7pt}

% \begin{tabular}{l|c|ccccccc}
% \toprule
% \multirow{2}{*}{\textbf{Method}} & \multirow{2}{*}{\textbf{Setting}} & \multicolumn{5}{c}{\textbf{Success Rate (\%)}} & \multirow{2}{*}{\textbf{FLOPs (\%)}} & \multirow{2}{*}{\textbf{Speedup}} \\ 
% \cmidrule(lr){3-7}
%  &  & \textbf{Spatial} & \textbf{Object} & \textbf{Goal} & \textbf{10} & \textbf{average} &  &  \\ 
% \midrule

% Baseline & FP16 & 57 & 75 & 60 & 19 & 52.75 & 100 & 1 \\ 
% QVLA & W4A8 & 54 & 73 & 58 & 16 & 50.25 & 12.625 & 1.47 \\ 
% QuantVLA & W4A8 & 47 & 72 & 53 & 12 & 46 & 16.125 & 1.325 \\ \hline % 区分不同技术路线的细线

% Eventful & Pruning & 52 & 70 & 57 & 15 & 48.5 & 56.38158 & 2.535 \\ 
% MaskVD & Pruning & 54 & 72 & 58 & 15 & 49.75 & 69.97732 & 1.765 \\ 
% SP-VLA & Pruning & 55 & 71 & 55 & 17 & 49.5 & 72.58 & 1.5025 \\ \hline % 区分不同技术路线的细线

% sVLA & W8A3 & 57 & 74 & 60 & 19 & 52.5 & 10.22321 & 8.777058 \\ 
% \bottomrule
% \end{tabular}
% \end{table*}

\subsection{Workloads}
To validate the effectiveness and deployability of SpecVLA, we evaluate our system using OpenVLA~\cite{kim2024openvla} and RDT~\cite{liu2024rdt} as base models across diverse robotic manipulation environments. Specifically, we deploy OpenVLA on the Franka Emika Panda arm and evaluate its performance within the LIBERO benchmark~\cite{liu2023libero}, which interfaces with real-time camera streams and robot proprioceptive states. We use four distinct task suites: LIBERO-Spatial, LIBERO-Object, LIBERO-Goal, and LIBERO-10, each providing 500 expert demonstrations across 10 tasks to stress-test SpecVLA’s generalization under varying spatial layouts, object distributions, and long-horizon planning. For the RDT model, we evaluate it on the ManiSkill platform~\cite{taomaniskill3}, which supports a diverse fleet of physical robot embodiments (humanoids, mobile manipulators, single-arm robots) as well as a wide range of tasks (table-top, drawing/cleaning, dextrous manipulation). We use three representative task suites: ManiSkill-PickCube, ManiSkill-PushCube, and ManiSkill-PlugCharger to evaluate the performance of RDT. 
% By evaluating SpecVLA on these standardized platforms, we demonstrate the operational viability of our speculative paradigm in diverse manipulation workloads.

\subsection{Algorithm Evaluation}

\textbf{Methodology.} 
We employ open-source implementations for inference and finetuning of the aforementioned VLAs, running on the PyTorch framework~\cite{paszke2019pytorch}. We implement the proposed Speculative VLA Inference and Action Verification algorithm (we will use the SpecVLA algorithm to represent it in the evaluation chapter) in Python and integrate it into the models' implementations. In the experiment, we use 0-bit, 4-bit, and 8-bit for sVLA.

\begin{figure}[t]
% \vspace{-0.1cm}
\centering
\includegraphics[width=\linewidth]{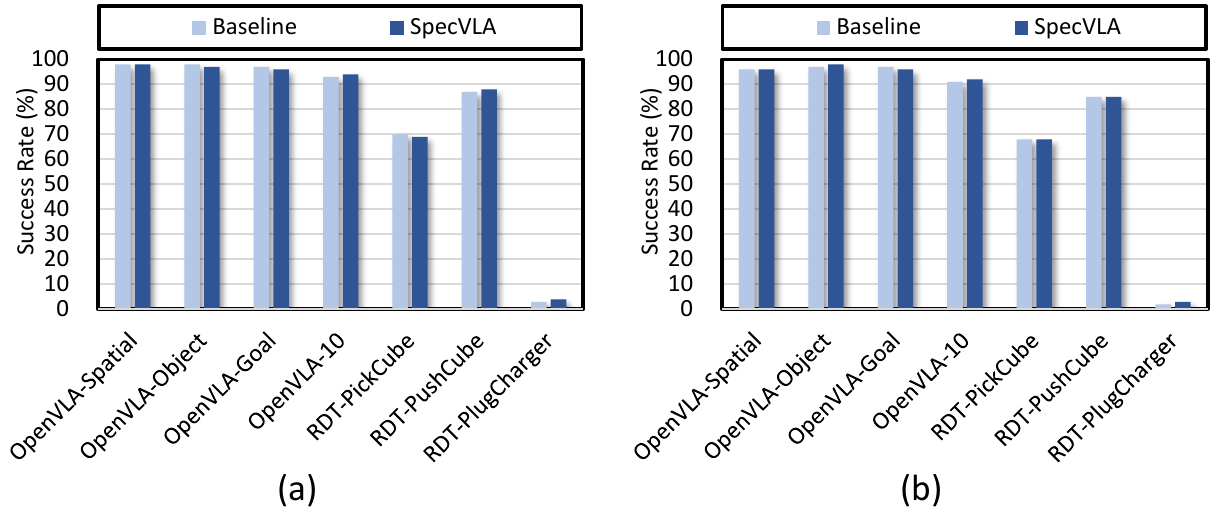}
\vspace{-0.8cm}
\caption{Model success rate results: (a) Performance under noise-free environments; (b) Performance with incorporated sensor noise.}
\vspace{-0.6cm}
\label{fig-eva-success-rate}
\end{figure}

\textbf{Success Rate.} 
We use success rate to measure the accuracy of each workload. Success rate refers to the proportion of successful tasks in the total tasks, which can be used to measure the model's ability to complete tasks. As illustrated in Fig.~\ref{fig-eva-success-rate}(a), SpecVLA achieves a negligible loss in success rate over the baselines, confirming its reliability in noise-free environments. Furthermore, to evaluate the system's robustness, we model and integrate sensor noise into our system by mimicking adverse physical conditions such as low-illumination and foggy scenarios based on prior methodologies~\cite{cao2023physics,dirndorfer2011model}. All subsequent evaluations are conducted under these non-ideal conditions to prioritize real-world applicability over laboratory-standard settings. As illustrated in Fig.~\ref{fig-eva-success-rate}(b), compared to the baseline models, SpecVLA maintains a $0.3\%$ performance lead despite these perturbations, thereby validating its robustness in real-world environments with compromised sensor precision. It is worth noting that the baseline RDT model achieves only a $2\%$ success rate on the PlugCharger task, which is due to its limited generalization capability in complex environments. In such a case, the success rate of SpecVLA is still $1\%$ higher than the baseline RDT model, which fully demonstrates that SpecVLA is effective in difficult tasks.
% We use success rate to measure the accuracy of each workload. Success rate refers to the proportion of successful tasks in the total tasks, which can be used to measure the model's ability to complete tasks. As illustrated in Fig.~\ref{fig-eva-success-rate}(a), compared to the baseline models, the SpecVLA algorithm exhibits average improvement in success rate of $0.6\%$. We notice that the baseline RDT model achieves only a $3\%$ success rate on the PlugCharger task, which is due to its limited generalization capability in complex environments. In contrast, the success rate of SpecVLA is still $1\%$ higher than the baseline, which fully demonstrates that SpecVLA is effective in difficult tasks. Furthermore, to evaluate the system's robustness, we model sensor noise and incorporate it into our system to investigate its impact on success rates. As illustrated in Fig.~\ref{fig-eva-success-rate}(b), SpecVLA sustains a stable success rate comparable to the baseline despite these perturbations, thereby validating its reliability in real-world environments with compromised sensor precision.
% SpecVLA maintains a $0.3\%$ performance lead; As illustrated in Fig.~\ref{fig-eva-success-rate}(b), compared to the baseline models, the SpecVLA algorithm exhibits an average improvement in success rate of $0.3\%$ despite these perturbations
%重点讲噪声实验而不是无噪声实验，篇幅调整一下。讲噪声的时候加一下我们模拟了真实环境中低光照等不理想的噪声，要像具体例子那样体现真实性，最好能引用一下相关论文

\begin{figure}[h]
% \vspace{-0.1cm}
\centering
%\vspace{-0.4cm}
\includegraphics[width=0.9\linewidth]{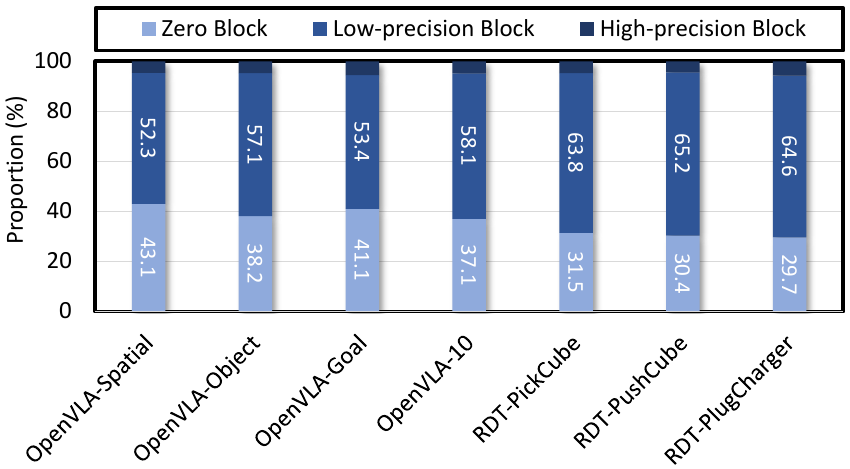}
\vspace{-0.5cm}
\caption{The proportion of zero block, low-precision block and high-precision block in the sVLA model.}
\vspace{-0.4cm}
\label{fig-eva-block-percentage}
\end{figure}

We also report the proportion of zero block (0-bit), low-precision block (4-bit), and high-precision block (8-bit) in sVLA models. Fig.~\ref{fig-eva-block-percentage} demonstrates that the differential-based hardware-friendly quantization scheme creates an average of $35.8\%$, $59.3\%$, and $4.9\%$ zero blocks, low-precision blocks, and high-precision blocks respectively. These results indicate that our proposed construction method can effectively reduce the computational complexity of the sVLA.

\textbf{Analysis of Speculation Efficiency.}
To assess the effectiveness of the speculative paradigm in SpecVLA, we conduct a quantitative analysis focusing on key metrics including the ratio of active/inactive state, rollback frequency, effective action length and path length, as summarized in Table~\ref{tab-eva-spec-analy}. First, we observe that the inactive states dominate the execution with an average ratio of $74.2\%$. This implies that in the vast majority of cycles, the robotic environment is sufficiently stable and safe. Second, the average rollback frequency remains as low as $9.5\%$. Moreover, to test the stability of SpecVLA, we select long-horizon tasks with complex layouts and diverse objects (called Complex Task in the table) from the benchmarks to measure its rollback frequency. The results show that SpecVLA stays well-bounded at $16.4\%$ even across the complex cases, which indicates that our speculative paradigm achieves a balance between minimal misprediction risk and high speculative gains. Third, with a speculative length of $8$, SpecVLA achieves an average accepted action length of $7.1$ per invocation. This represents a $3.5\times$ reduction in expensive VLA invocations compared to the non-speculative baseline, effectively minimizing the end-to-end latency. Fourth, compared to the non-speculative baseline, the path length required for SpecVLA to complete tasks remains nearly unchanged ($\pm2\%$), indicating that path deviations or re-execution overhead introduced by speculation are negligible. In contrast, MoLe-VLA expands the execution path length by $14.8\%$ for the same tasks. This efficiency loss occurs because such layer-skipping approaches inherently sacrifice the model's expressive depth, frequently synthesizing suboptimal actions that prolong physical execution.
% reports the total path length of task completion and the effective action length per VLA invocation. Our results show that, compared to a non-speculative baseline, the path length of task completion remains nearly unchanged ($\pm2\%$), indicating that VLA reinvocation cost caused by speculation is negligible.
%加一个表，对应这段文字和dadu-corki的action length

\begin{table}[h]
\centering
\small
\renewcommand{\arraystretch}{1.3}
\setlength{\tabcolsep}{7pt}
\resizebox{\linewidth}{!}{
\begin{tabular}{l|c|cccc|c|c}
\toprule
\multirow{2}{*}{\textbf{Method}} & \multirow{2}{*}{\textbf{Setting}} & \multicolumn{4}{c|}{\textbf{Success Rate (\%)}} & \multirow{2}{*}{\textbf{FLOPs (\%)}} & \multirow{2}{*}{\textbf{Speedup}} \\ 
\cmidrule(lr){3-6}
 &  & \textbf{Spatial} & \textbf{Object} & \textbf{Goal} & \textbf{10} &  &  \\ 
\midrule

OpenVLA & FP16 (GPU) & 98 & 98 & 97 & 93 & 100 & 1 \\ \hline % 区分不同技术路线的细线

QVLA~\cite{xu2026qvla} & W4A8 (GPU) & 95 & 96 & 94 & 90 & 12.63 & 1.47 \\ 
QuantVLA~\cite{zhang2026quantvla} & W4A8 (GPU) & 89 & 94 & 90 & 87 & 16.13 & 1.32 \\ \hline % 区分不同技术路线的细线

Eventful~\cite{dutson2023eventful} & Pruning (GPU) & 93 & 92 & 94 & 89 & 56.38 & 2.54 \\ 
MaskVD~\cite{sarkar2025maskvd} & Pruning (GPU) & 94 & 95 & 93 & 89 & 69.98 & 1.76 \\ 
SP-VLA~\cite{li2025sp} & Pruning (GPU) & 95 & 94 & 93 & 90 & 72.58 & 1.51 \\ \hline % 区分不同技术路线的细线

Our Method & W8A3 (NPU) & 98 & 97 & 97 & 93 & 10.22 & 8.78 \\ 
\bottomrule
\end{tabular}
}
\caption{Quantitative Comparison of Differential-based, Hardware-friendly Quantization with Existing Quantization Methods and Pruning Methods on LIBERO Benchmark.}
% \vspace{-0.8cm}
\vspace{-0.8cm}
\label{tab-eva-svla-analy}
\end{table}

\textbf{Analysis of the Differential-based, Hardware-friendly Quantization.} To demonstrate the efficacy and necessity of our proposed differential-based, hardware-friendly quantization scheme, we conduct a multi-dimensional analysis against state-of-the-art model compression methods using the OpenVLA model on the LIBERO benchmark. The baselines encompass representative quantization methods (QVLA~\cite{xu2026qvla} and QuantVLA~\cite{zhang2026quantvla}) and pruning approaches (Eventful Transformer~\cite{dutson2023eventful}, MaskVD~\cite{sarkar2025maskvd}, and SP-VLA~\cite{li2025sp}). The quantitative results are summarized in Table~\ref{tab-eva-svla-analy}.

Compared to SOTA quantization baselines, our scheme achieves superior task success rates and execution efficiency. Specifically, the average success rate of our method outperforms QVLA and QuantVLA by $2.50\%$ and $6.25\%$, respectively. This performance gap stems from a fundamental limitation of prior methods: by statically exploring quantization potential based solely on intra-model feature distributions, they neglect the temporal continuity inherent in embodied AI workflows, leading to degraded task fidelity under ultra-low bit-width constraints. In terms of computational efficiency, our method achieves a $1.23\times$ and $1.58\times$ reduction in theoretical FLOPs compared to QVLA and QuantVLA, respectively. Critically, our method delivers an $8.78\times$ end-to-end speedup over the uncompressed OpenVLA baseline, and achieves $5.97\times$ and $6.62\times$ acceleration over QVLA and QuantVLA. These results demonstrate that our custom Robotic-Specific Hardware effectively unleashes the acceleration potential of our quantization scheme.

Table~\ref{tab-eva-svla-analy} also highlights the comparative advantages of our scheme over mainstream pruning strategies. Our method improves the average task success rate by $4.25\%$, $3.50\%$, and $3.25\%$ compared to the Eventful Transformer, MaskVD, and SP-VLA, respectively. This improvement of accuracy is primarily attributed to our design principle that avoids aggressively discarding critical spatial information. Besides, our method compresses the theoretical FLOPs to merely $10.22\%$ of the OpenVLA model, whereas Eventful Transformer, MaskVD, and SP-VLA retain $56.38\%$, $69.98\%$, and $72.58\%$ of the baseline computations. Furthermore, the end-to-end execution results reveal that our method running on the Robotic-Specific Hardware achieves $8.78\times$, $3.46\times$, $4.97\times$, and $5.84\times$ speedup compared to the OpenVLA model, Eventful Transformer, MaskVD, and SP-VLA deployed on the GPU. These comprehensive evaluations conclusively validate the hardware-software co-design advantages of our framework in real-time embodied manipulation scenarios.

\subsection{Architecture Evaluation}\label{ssect:exp-system}
\textbf{Methodology.} 
To evaluate the performance of the SpecVLA heterogeneous architecture, we develop a cycle-level simulator to collect the latency statistics of robotic-specific hardware module for each workload. The simulator is integrated with Ramulator~\cite{kim2015ramulator} for DRAM timing. We make every effort to ensure the accuracy of the simulator by following the widely adopted open-source simulator, Scale-Sim~\cite{samajdar2018scale}. Moreover, we implement the proposed robotic-specific hardware module in Verilog and synthesize it by Synopsys Design Compiler to get the chip area and total power under 28nm technology with a frequency of 500MHz. This synthesis process generates a comprehensive report containing the gate-level netlist, timing information, and area breakdown of various components within the robotic-specific hardware module. We also employ CACTI~\cite{balasubramonian2017cacti} to derive the energy and area of on-chip buffers based on parameters such as bus width, size, and the number of reads/writes. The technology nodes used in robotic-specific hardware module are calibrated using the ScaleTool~\cite{stillmaker2017scaling}. For GPU, we use the CUDA event API to measure its actual latency and nvidia-smi to measure its power during runtime.

%Additionally, we compare with a variant of the AQuant architecture, denoted as AQuant-full, which operates with full-precision ViT models on the AQuant architecture. 

% We implement the SpecVLA during the evaluation, namely Server-SPVLA and Edge-SPVLA. Server-SPVLA consists of a server NPU and a representative server GPU NVIDIA A100 80GB PCIE. We will compare them separately with NVIDIA A100 and NVIDIA AGX Xavier to evaluate the performance of our proposed SPVLA in different scenarios. For a fair comparison with GPU, we scale the number of computing cores of our SPVLA to ensure the same area budget. This is achieved by leveraging the PyTorch framework to execute VLA on Xavier AGX and A100, recording their execution time, and then scaling the time based on the ratio of the number of cores in the GPU to those in the SPVLA system.
We implement SpecVLA in our evaluation, which consists of a robotic-specific hardware module and a representative server GPU——NVIDIA A100 80GB PCIe. Then we compare it against several platforms, including Intel(R) Xeon(R) Gold 6226R CPU, NVIDIA A100 GPU, one related accelerator Dadu-Corki-ADAP~\cite{DBLP:conf/isca/HuangH0Y0M00L0G25} (with adaptive trajectory length). For Dadu-Corki-ADAP, we implement its waypoint extraction and identification mechanism and evaluate it on our benchmarks to enable adaptive trajectory lengths. In addition to these hardware baselines, we evaluate SpecVLA against recent algorithm-only acceleration works, including MoLe-VLA~\cite{zhang2026mole}, VLA-Cache~\cite{xu2026vla}, Spec-VLA~\cite{wang2025spec} (denoted as prior work~\cite{wang2025spec} to avoid confusion), and HeiSD~\cite{zheng2026heisd}, to validate the comprehensive advantages of our hardware-software co-design.
% For a fair comparison, we scale up the number of computing cores of SpecVLA to ensure an identical area budget.

\begin{figure}[t]
% \vspace{-0.1cm}
\centering
\includegraphics[width=0.9\linewidth]{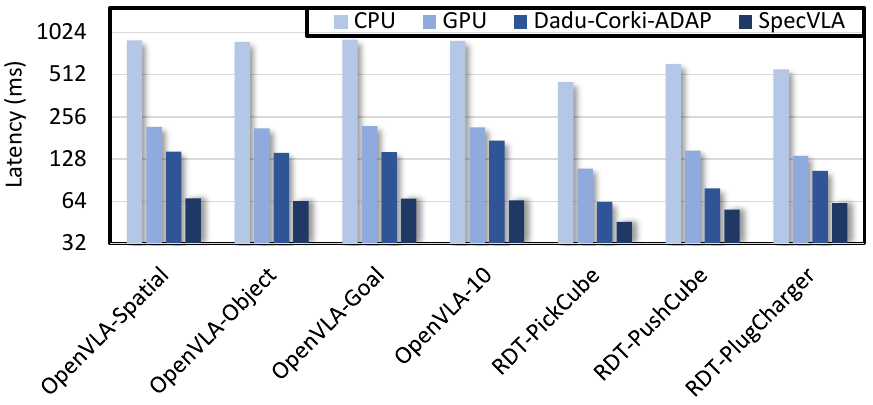}
\vspace{-0.4cm}
\caption{Per-action latency results.}
\vspace{-0.3cm}
\label{fig-eva-speedup}
\end{figure}

% \begin{figure*}[h]
% \centering
% \includegraphics[width=\linewidth]{figures/latency_breakdown3.pdf}
% % \vspace{-0.4cm}
% \caption{\textcolor{blue}{Latency breakdown of SpecVLA in three dataflow modes.}}
% \label{fig-eva-cyclebreakdown}
% \end{figure*}

\begin{table}[t]
\centering
\small
\renewcommand{\arraystretch}{1.3}
\setlength{\tabcolsep}{7pt} 

% 定义 8 列：前两列后有竖线，Success Rate 内部 4 列无竖线，最后两列之间有竖线
\resizebox{\linewidth}{!}{
\begin{tabular}{l|c|cccc|c|c}
\toprule
\multirow{2}{*}{\textbf{Method}} & \multirow{2}{*}{\textbf{Setting}} & \multicolumn{4}{c|}{\textbf{Success Rate (\%)}} & \multirow{2}{*}{\textbf{Speedup}} & \multirow{2}{*}{\textbf{Energy Efficiency}} \\ 
\cline{3-6} % 横线仅覆盖 Success Rate 对应的第 3 到 6 列
 &  & \textbf{Spatial} & \textbf{Object} & \textbf{Goal} & \textbf{10} &  &  \\ 
\midrule

OpenVLA & FP16 (GPU) & 98 & 98 & 97 & 93 & 1 & 1 \\ \hline
MoLe-VLA~\cite{zhang2026mole} & Skipping Layer (GPU) & 97 & 98 & 96 & 92 & 1.98 & 1.98 \\ \hline
VLA-Cache~\cite{xu2026vla} & Token Reusing (GPU) & 97 & 96 & 96 & 92 & 1.26 & 1.26 \\ \hline
Spec-VLA~\cite{wang2025spec} & Speculation (GPU) & 94 & 94 & 96 & 87 & 1.32 & 1.32 \\ \hline
HeiSD~\cite{zheng2026heisd} & Speculation (GPU+CPU) & 94 & 98 & 92 & 86 & 2.04 & 1.69 \\ \hline
SpecVLA & Speculation (GPU+NPU) & 98 & 97 & 96 & 94 & 3.28\ & 2.29\ \\ 
\bottomrule
\end{tabular}}
\caption{Quantitative Comparison of SpecVLA with Existing Algorithm-only Acceleration Works on LIBERO Benchmark.}
% \vspace{-0.8cm}
\vspace{-1.0cm}
\label{tab-eva-specwork-analy}
\end{table}

\textbf{Latency.} 
Fig.~\ref{fig-eva-speedup} showcases the performance of the SpecVLA, which includes the per-action latency of Intel CPU running the full-precision VLAs (marked as CPU), A100 GPU running the full-precision VLAs (marked as GPU), and Dadu-Corki-ADAP. On average, the SpecVLA achieve $12.1\times$, $2.9\times$, and $1.9\times$ speedup over CPU, GPU, and Dadu-Corki-ADAP, respectively. Furthermore, Table~\ref{tab-eva-specwork-analy} presents a quantitative comparison between SpecVLA and state-of-the-art algorithmic acceleration methods. The results demonstrate that our framework delivers $1.66\times$, $2.59\times$, $2.48\times$, and $1.61\times$ speedups over MoLe-VLA, VLA-Cache, prior work~\cite{wang2025spec}, and HeiSD, respectively. The performance improvement stems from two factors. First, the SpecVLA algorithm reduces the number of VLA invocations by increasing the action length. As reported in Table~\ref{tab-eva-spec-analy}, SpecVLA achieves an average action length of $7.1$, resulting in less model invocations. Second, the speculative dataflow enables system-level parallelization of the VLA, sVLA, and action execution, thereby hiding part of the total latency. To verify whether the SpecVLA system meets the real-time requirement, we measure the average latency required for SpecVLA to generate an action. We find that SpecVLA generates one action at an average interval of about $61$ms, which is sufficient to meet the real-time requirement ($<100$ms)~\cite{liu2026draco}. %This result is obtained by the contribution of both algorithm and hardware, where the SpecVLA algorithm increases the average prediction length from 2 to 7.2 and the SpecVLA system reduced the average inference latency by 0.42s.
\begin{table}[h]
\vspace{-0.2cm}
\centering
\small
\resizebox{\columnwidth}{!}{%
\begin{tabular}{c | c | c | c}
\toprule
\textbf{Modules} & \textbf{Components} & \textbf{Area($mm^2$)} & \textbf{Power($mW$)} \\
\midrule
\multirow{2}{*}{PE Array} & 8192 4-bit$\times$16-bit PE Units & \multirow{2}{*}{1.58} & \multirow{2}{*}{495.06} \\
 & 680KB On-chip Buffer & & \\
\midrule
\multirow{2}{*}{State Predictor} & 7 Subtractors & \multirow{2}{*}{0.03} & \multirow{2}{*}{13.42} \\
 & 3-stage Adder Trees & & \\
\midrule
\multirow{4}{*}{Preprocessing Module} & 8$\times$32 Subtractors & \multirow{4}{*}{0.32} & \multirow{4}{*}{59.26} \\
 & 8$\times$32 Multipliers & & \\
 & 5-stage Adder Trees & & \\
 & 8KB Block Buffer & & \\
% \cmidrule{2-4} %
%  & 8KB Block Buffer & 0.04 & 4.98 \\
\midrule
\multicolumn{2}{c|}{Total} & 1.93 & 567.74 \\
\bottomrule
\end{tabular}
}
\caption{Area of the Robotic-specific Hardware Module.}
% \vspace{-0.6cm}
\vspace{-0.8cm}
\label{tab-eva-npu-area}
\end{table}

\textbf{Hardware Overhead and Area.} 
Table~\ref{tab-eva-npu-area} provides a comprehensive breakdown of design parameters, area, and power of the robotic-specific hardware module in the SpecVLA architecture. The PE array of the robotic-specific hardware module comprises $8\times 32\times 32$ 4-bit$\times$16-bit PE units. We also evaluate the costs of the preprocessing module and the environmental state predictor. All above modules together occupy $1.93mm^2$ of the total area and $567.74$mW power consumption.

\begin{figure}[h]
\vspace{-0.2cm}
\centering
%\vspace{-0.4cm}
\includegraphics[width=0.9\linewidth]{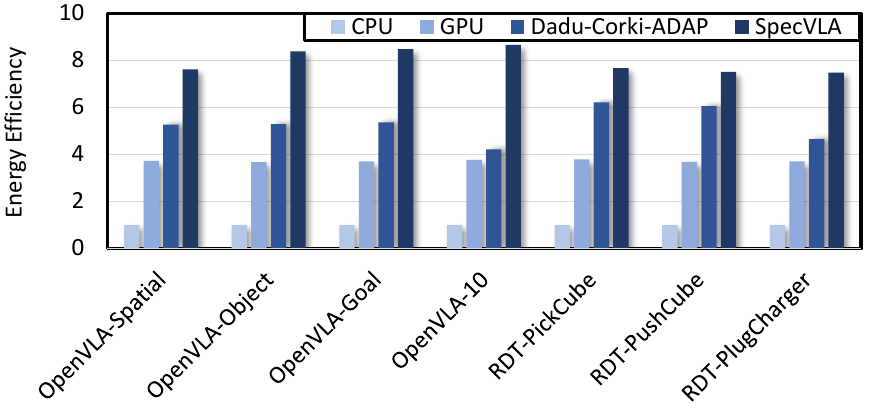}
\vspace{-0.4cm}
\caption{Energy efficiency results.}
\vspace{-0.4cm}
\label{fig-eva-energy}
\end{figure}

\textbf{Energy Efficiency.} 
The energy efficiency outcomes are depicted in Fig.~\ref{fig-eva-energy} and Table~\ref{tab-eva-specwork-analy}. The SpecVLA delivers remarkable energy efficiency, which surpasses CPU, GPU, Dadu-Corki-ADAP, MoLe-VLA, VLA-Cache, prior work~\cite{wang2025spec}, HeiSD by $8.0\times$, $2.1\times$, $1.5\times$, $1.2\times$, $1.8\times$, $1.7\times$, $1.4\times$, respectively. These substantial energy savings stem from two primary factors. First, our proposed framework significantly reduces both the total computations and the end-to-end latency. Second, the customized robotic-specific hardware inherently delivers vastly superior energy efficiency compared to power-intensive general-purpose GPUs.
% \textbf{Energy Efficiency.} 
% The energy efficiency outcomes are depicted in Fig.~\ref{fig-eva-energy}. The SpecVLA delivers remarkable energy efficiency, which surpasses CPU, GPU, and Dadu-Corki-ADAP by $8.0\times$, $2.1\times$, $1.5\times$, respectively. These substantial savings in energy consumption come from the reduction in computations and total latency.

\begin{figure}[h]
\centering
\includegraphics[width=0.9\linewidth]{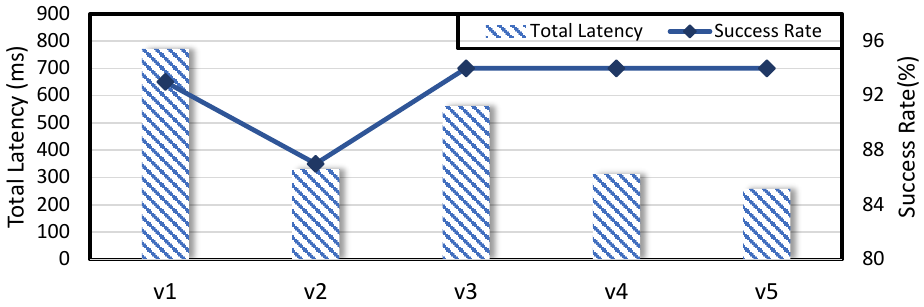}
\vspace{-0.4cm}
\caption{Ablation study of software and hardware contributions. Configurations v1–v5 correspond to: (1) Traditional VLA on GPU;
(2) sequential SpecVLA (speculative action length = 8) on GPU;
(3) sequential SpecVLA (speculative action length = 8) with verification on GPU;
(4) SpecVLA with speculative dataflow on GPU;
(5) SpecVLA with speculative dataflow and the robotic-specific hardware module.}
\vspace{-0.5cm}
\label{fig-eva-ablation-study}
\end{figure}

\textbf{Ablation Study of Software and Hardware Contributions.} 
We add ablations with five configurations in Fig.~\ref{fig-eva-ablation-study}. First, comparing v2 with v1 isolates the benefit of longer action length. Second, comparing v3 with v2 quantifies the impact of the verification scheme, which improves task success rate by approximately $7\%$, suggesting its necessity for maintaining correctness under speculative execution. Third, v5 and v4 comparison shows robotic-specific accelerator further reduces end-to-end latency by $20.6\%$, quantifying the portion of speedup attributable to the custom hardware module. Finally, speculative dataflow (v4 vs. v3) contributes an additional $79.8\%$ latency reduction by overlapping VLA prediction and sVLA verification.

\begin{figure*}[h]
\centering
\includegraphics[width=\linewidth]{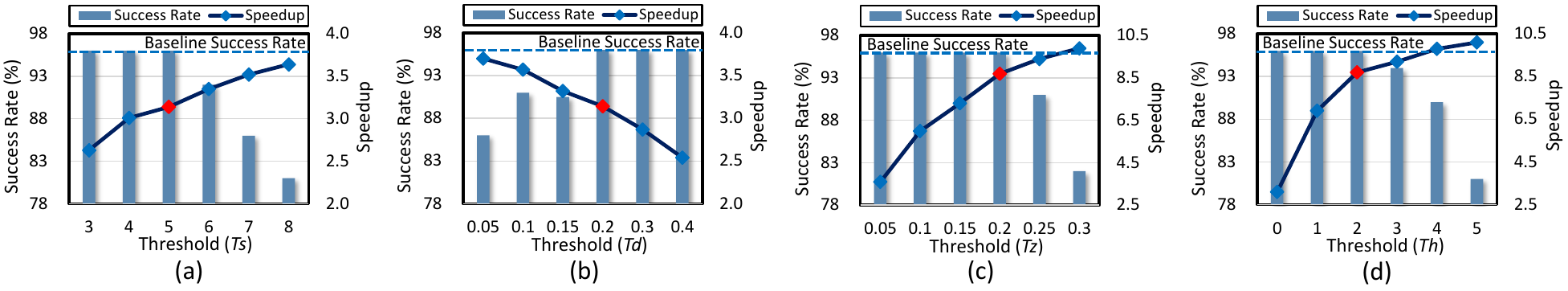}
\vspace{-0.8cm}
\caption{Design space exploration.}
% \vspace{-0.4cm}
\vspace{-0.4cm}
\label{fig-eva-bias-exploration}
\end{figure*}

\subsection{Design Exploration}
% Need：\usepackage{subcaption}

In this section, we explore four thresholds $T_s$, $T_d$, $T_z$, and $T_h$ by making a tradeoff between the success rate and speedup. Our threshold selection prioritizes robustness over global optimality. Instead of jointly searching all $6^4=1296$ combinations, we adopt a heuristic-based one-dimensional sweep strategy: when tuning one threshold, others are fixed to the highest-accuracy configuration, and we select the largest speedup before a success-rate cliff drop to ensure that performance improvements do not come at the cost of reliability.

\textbf{Exploration of Threshold $T_s$.} The key of the SpecVLA algorithm is to verify the acceptability of Predicted Actions, which is controlled by the threshold $T_s$. Therefore, $T_s$ directly determines the acceptability of Predicted Actions, which in turn affects the success rate and speedup. Specifically, a larger $T_s$ means that more Predicted Actions are verified as accurate, which results in higher speedup but lower success rate. To explore the impact of $T_s$, we vary $T_s$ from $3$ to $8$ and measure the success rate and speedup of OpenVLA model on LIBERO dataset. As illustrated in Fig.~\ref{fig-eva-bias-exploration}(a), when $T_s$ increases from $3$ to $5$, more Predicted Actions are identified as accurate, leading to higher speedup. But when we keep increasing $T_s$, the success rate drops severely. As a result, we set $T_s$ as $5$ to ensure both the success rate and performance.

\textbf{Exploration of Threshold $T_d$.} In environmental state prediction, the threshold $T_d$ determines whether the robotic arm is close to the objects in the environment, thus affecting the prediction result. Specifically, a larger $T_d$ means that more objects are considered to be close to the robotic arm, which results in the robotic arm being in an active state more frequently. This is good for maintaining the success rate, but not friendly for improving speedup. To explore the impact of $T_d$, we vary $T_d$ from $0.05$ to $0.4$ and observe the success rate and speedup of OpenVLA model on LIBERO dataset. As illustrated in Fig.~\ref{fig-eva-bias-exploration}(b), when $T_d$ decreases from $0.4$ to $0.2$, fewer states are identified as active states, leading to higher speedup. But when we keep decreasing $T_d$, the success rate drops severely. As a result, we set $T_d$ as $0.2$ to ensure both the success rate and performance.

\textbf{Exploration of Thresholds $T_z$ and $T_h$.} In differential-based hardware-friendly quantization, $T_z$ and $T_h$ determine the precision of each residual block, thus affecting the computational overhead and accuracy of the sVLA model. Specifically, a larger $T_z$/$T_h$ means that more blocks are quantized to 0-bit/4-bit, which results in lower computational overhead for sVLA. This helps reduce the cost of action verification, but is not conducive to ensuring the accuracy of verification. To explore the impact of $T_z$, we vary $T_z$ from $0.05$ to $0.3$ and observe the success rate and speedup of sVLA model constructed from OpenVLA on robotic-specific hardware module. As illustrated in Fig.~\ref{fig-eva-bias-exploration}(c), when $T_z$ increases from $0.05$ to $0.2$, more blocks are quantized as 0-bit, leading to higher speedup. But when we keep increasing $T_z$, the success rate drops severely. As a result, we set $T_z$ as $0.2$ to ensure both the success rate and performance. Similarly, we do the same exploration for $T_h$ and set $T_h$ to $2$.

\begin{figure}[h]
\vspace{-0.2cm}
\centering
\includegraphics[width=0.9\linewidth]{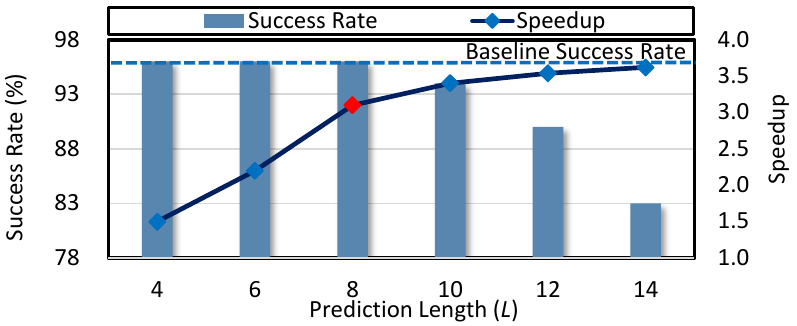}
\vspace{-0.5cm}
\caption{Exploration of prediction length.}
\vspace{-0.3cm}
\label{fig-eva-exploration-prediction-length}
\end{figure}

\textbf{Exploration of prediction length.} To study the effect of the number of actions generated by VLA model (prediction length, denoted as $L$), we add a sensitivity analysis. Specifically, we vary $L$ from $4$ to $14$ and evaluate its impact on both speedup and success rate. As shown in Fig.~\ref{fig-eva-exploration-prediction-length}, when $L$ increases from $4$ to $8$, the success rate remains stable. This is because the predicted action sequence is still within the reliable prediction horizon of the VLA model. Meanwhile, increasing $L$ better amortizes the expensive VLA inference cost across more actions, leading to an improvement in speedup. However, when $L$ further increases beyond $8$, the success rate starts to decrease, while the speedup improvement becomes marginal. This is because a long predicted sequence is more likely to mismatch the latest physical environment, leading to more frequent verification failures, rollbacks, and VLA reinvocations, thus reducing the effective benefit of longer prediction. In addition, we observe that overly long action sequences also reduce the success rate. Based on this sensitivity analysis, we set $L$ to $8$ in our all experiments, as it provides the best trade-off between maintaining success rate and improving speedup.
% To further evaluate robustness and generalization of the selected thresholds, we directly apply them to highly dynamic scenarios held out from LIBERO and ManiSkill benchmarks without per-dataset tuning. As illustrated in Fig.~\ref{fig-eva-worstcase-exploration}, SpecVLA consistently achieves a $2.3\times$ speedup over GPU with negligible success rate degradation ($\leq0.5\%$). These results highlight that the performance gains and reliability of SpecVLA are not confined to average-case scenarios but extend to challenging environments, demonstrating its broad applicability.} 
% \vspace{-0.8cm}
% \vspace{-0.2cm}
\section{Related Works}

Traditional robots typically depend on optimization-based algorithms for decision-making and task planning~\cite{hao2024orianna,zhang2014loam}. In contrast, emerging robotic applications~\cite{li2023vision,wake2024gpt,hu2023toward,firoozi2023foundation} use Large Language Models (LLMs) to control robots for tasks such as object manipulation, task planning, and navigation, demonstrating capabilities far superior to traditional algorithms. As robots are increasingly treated as the next generation of computing platforms, many accelerators have been designed for motion planning~\cite{bakhshalipour2022racod,huang2024moped,hao2023blitzcrank,hsiao2023vapr,murray2019programmable,murray2016microarchitecture,lian2018dadu}, control~\cite{lian2017dadu, yang2023dadu,neuman2023roboshape,sacks2018robox,gac2012fpga,shao2018towards}, and navigation~\cite{yu2020building,krishnan2022automatic,lee2024spade}. However, these accelerators only focus on traditional optimization-based algorithms. In contrast, our work focuses on combining innovations in both algorithms and architecture to accelerate deep learning-based algorithms, distinguishing our work from previous research.

Our work is related to broad themes of temporal redundancy pruning~\cite{8574537, duan2022differential, zhang2021differentiable, dutson2023eventful, buckler2018eva2, sarkar2025maskvd, li2025sp, xu2026vla}, dynamic layer-skipping~\cite{yue2024deer, zhang2026mole}, and speculative execution~\cite{wang2025spec, zheng2026heisd, gagrani2024speculative, miao2024specinfer, yang2023inference, yang2023predictive, sun2023spectr, zhang2024draft}. While redundancy-based methods prune spatial information and layer-skipping approaches truncate model depth to reduce computations, they often compromise representation accuracy and yield suboptimal action that lengthen task execution. In contrast, SpecVLA focuses on extending the prediction length to amortize inference overhead without sacrificing expressive capacity of VLA models, thereby reducing computation while sustaining high task execution efficiency. Furthermore, distinct from existing speculative frameworks such as Spec-VLA~\cite{wang2025spec} and HeiSD~\cite{zheng2026heisd}, SpecVLA introduces a novel parallel prediction and serial verification paradigm. Our framework sequentially validates each action against real-time environmental feedback. This closed-loop responsiveness ensures execution accuracy during abrupt physical mutations (such as abrupt changes in direction), yielding a significant improvement in success rate over tranditional speculation works. Moreover, SpecVLA introduces a speculative dataflow on a heterogeneous architecture to enable hardware-level execution overlap between prediction and verification. This concurrent execution unlocks ultra-low latency and superior energy efficiency compared to conventional serial execution modes.
\vspace{-0.1cm}
\section{Conclusion}
% \vspace{-0.1cm}
Current VLAs exhibit excessive end-to-end latency, which is insufficient to meet the requirements of practical applications. This paper introduces SpecVLA, an algorithm-architecture co-design framework poised to reduce the end-to-end latency of the VLA model. The key idea of SpecVLA is to adaptively verify the predicted actions based on the environmental state. Experiments show that SpecVLA can reduce the end-to-end latency of the VLA models significantly while maintaining success rate, thus enabling real-time robotic manipulation in practical applications.
\bibliographystyle{plain}
\bibliography{sample-base}

\end{document}